\documentclass{article}

\usepackage{microtype}
\usepackage{graphicx}
\usepackage{booktabs}
\usepackage{hyperref}

\let\origaddcontentsline\addcontentsline
\usepackage[preprint]{icml2027}

\usepackage{amsmath}
\usepackage{amssymb}
\usepackage{mathtools}
\usepackage{amsthm}
\usepackage{enumitem}
\usepackage{longtable}
\usepackage{array}

\theoremstyle{plain}
\newtheorem{theorem}{Theorem}
\newtheorem{proposition}[theorem]{Proposition}
\newtheorem{corollary}[theorem]{Corollary}
\newtheorem{lemma}[theorem]{Lemma}
\newtheorem{definition}[theorem]{Definition}
\newtheorem*{hypothesis}{Hypothesis}
\theoremstyle{remark}
\newtheorem{remark}[theorem]{Remark}
\theoremstyle{plain}

\newcommand{\R}{\mathbb{R}}
\newcommand{\Z}{\mathbb{Z}}
\newcommand{\C}{\mathbb{C}}
\newcommand{\Acc}{\operatorname{Acc}}

\icmltitlerunning{Grokking as a Subspace Search}

\begin{document}

\twocolumn[
  \icmltitle{Grokking as a Subspace Search:\\ Computing the Destination Before Training}

  \begin{icmlauthorlist}
    \icmlauthor{Sunjae Jeong}{mob,khu}
    \icmlauthor{Jaeseung Kim}{mob}
    \icmlauthor{Hyoseok Hwang}{khu}
  \end{icmlauthorlist}

  \icmlaffiliation{mob}{MOBILTECH Co., Ltd.}
  \icmlaffiliation{khu}{Kyung Hee University}

  \icmlcorrespondingauthor{Hyoseok Hwang}{hyoseok@khu.ac.kr}

  \icmlkeywords{Grokking, Representation Theory, Learning Dynamics, Group Tasks}

  \vskip 0.3in
]

\printAffiliationsAndNotice{}

\begin{abstract}
Grokking is the phenomenon in which a neural network generalizes long after it has memorized its training data, and no explanation of the delay has determined before training where the network is heading. We show that on group tasks this destination can be computed from the task itself. The symmetry of the task fixes one family of candidate subspaces, and two requirements of the task select from it the subspace that we call the target subspace. This lets us test the hypothesis that the delay is a search for the target subspace, and that the progress of this search sets when the network generalizes. Restricted to the target subspace, the network generalizes in $150$ steps instead of $64{,}050$, before it has even fit its training set. With the target subspace ablated, it stops at an accuracy ceiling that we compute before training and that no trained network exceeds. In vanilla training, the network moves toward the target subspace during the delay, and moving it there during training brings generalization earlier. On every group task that we tested, grokking is a search for the target subspace, and the delay is the time that this search takes.
\end{abstract}

\section{Introduction}
\label{sec:intro}

\begin{figure}[t]
\centering
\includegraphics[width=0.89\columnwidth]{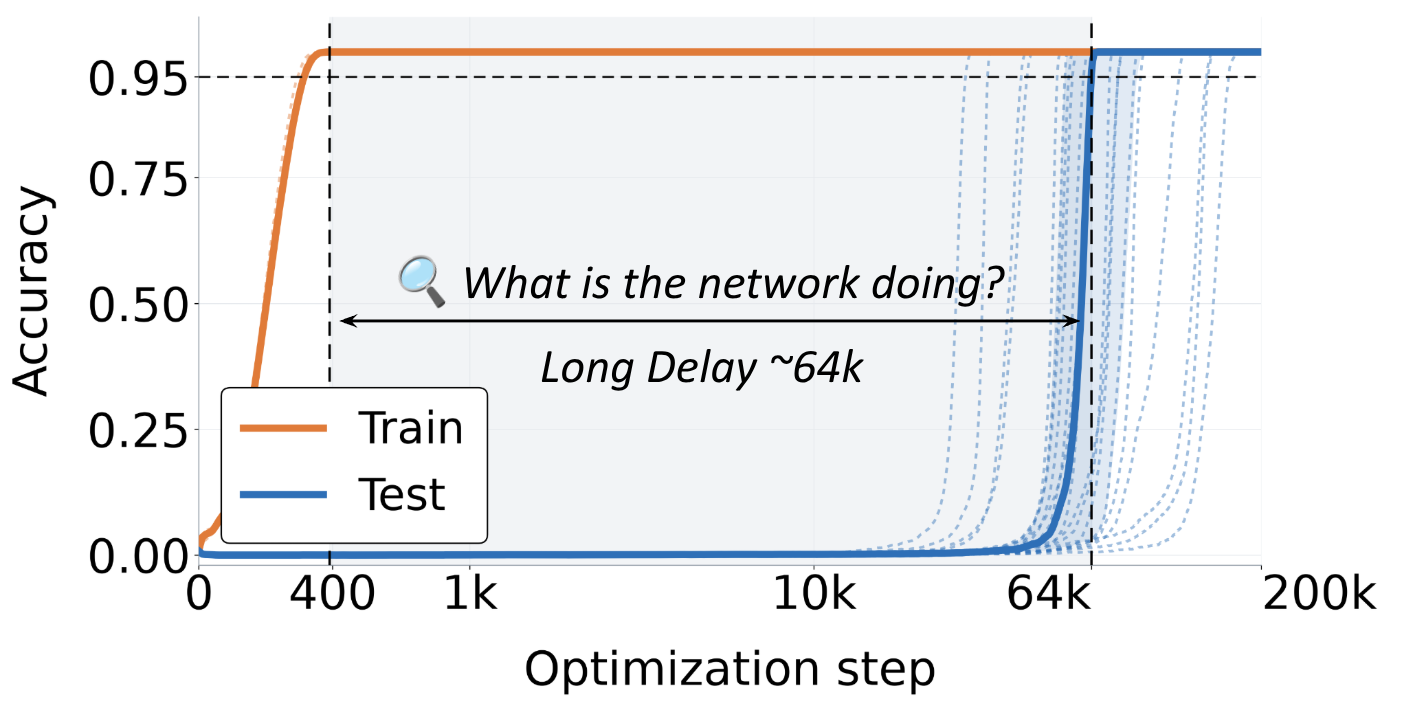}
\caption{\textbf{Grokking and the delay.} Twenty seeds of $S_5$ multiplication. Thick lines are medians.}
\label{fig:delay}
\end{figure}

Neural networks trained on small algorithmic tasks can generalize long after they have memorized their training data. \citet{power2022grokking} first observed this behavior and named it \emph{grokking}. In the typical case of Figure~\ref{fig:delay}, the network memorizes every training example within a few hundred steps, while its test accuracy stays at chance for tens of thousands of steps and then jumps abruptly to perfect generalization. In between lies a long delay in which nothing appears to happen. Grokking has been studied intensively since its discovery, yet the most basic question that it raises, what the network is doing during the delay, is still open.

\begin{figure*}[t]
\centering
\includegraphics[width=0.425\textwidth]{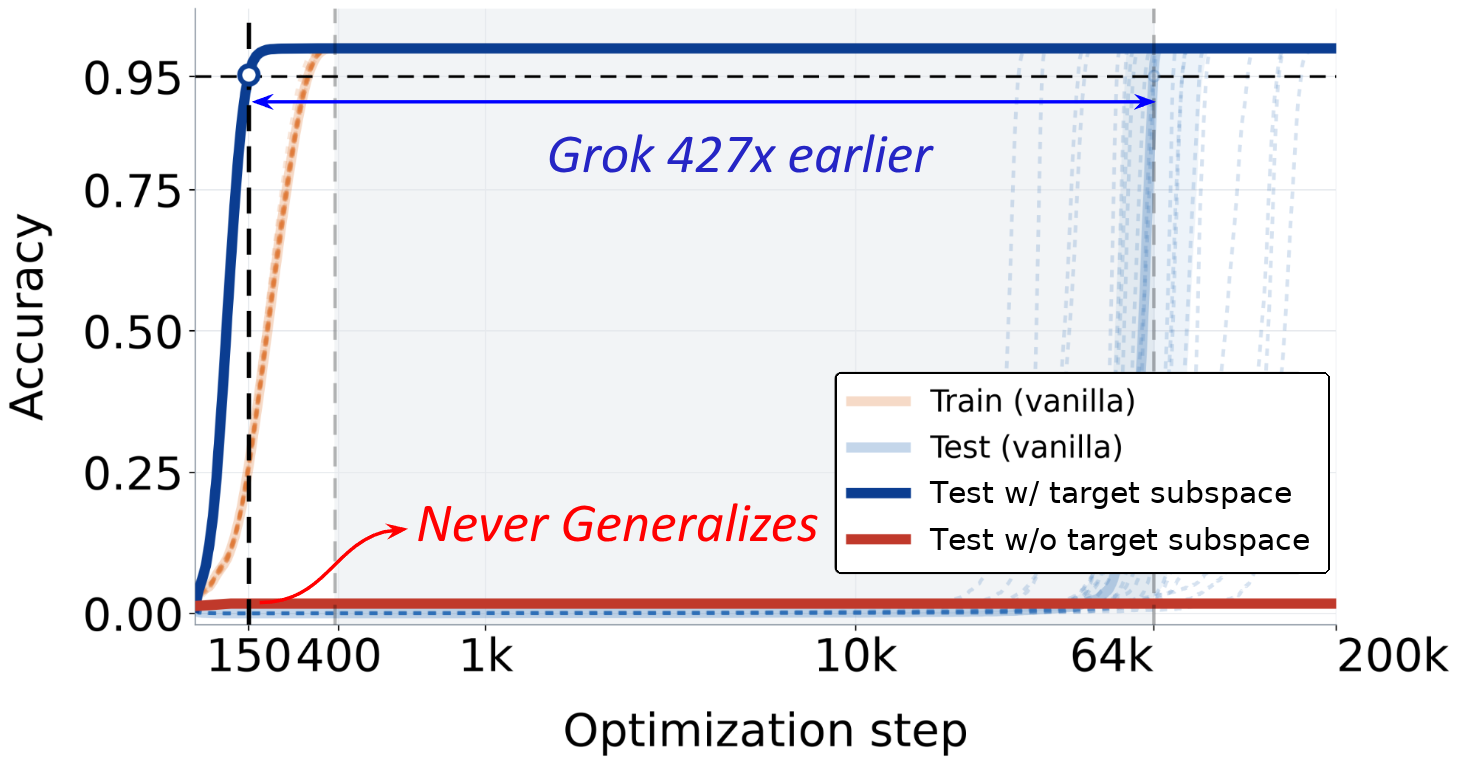}\hspace{0.05\textwidth}%
\includegraphics[width=0.425\textwidth]{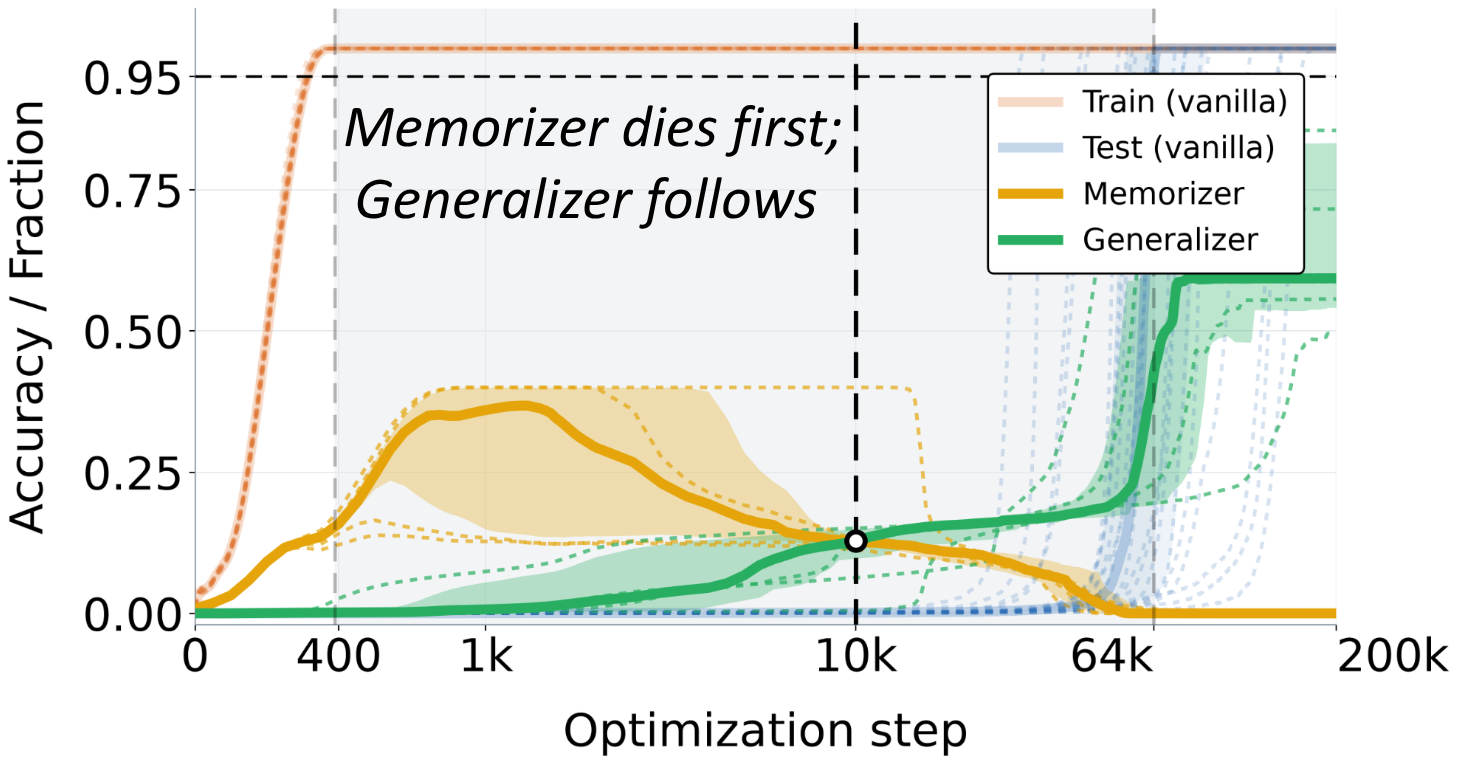}
\caption{\textbf{Restricting to the target subspace, ablating it, and observing vanilla training.} \textbf{Left.} Test accuracy with the network restricted to the target subspace and with the target subspace ablated. Faint curves are the vanilla runs of Figure~\ref{fig:delay}. \textbf{Right.} The memorizing and generalizing components of the logits in the vanilla runs, the parts that the irreps of the group do not and do explain.}
\label{fig:anatomy}
\end{figure*}

Prior work has given three kinds of answers. Reverse engineering shows that networks that have grokked implement structured algorithms~\citep{nanda2023progress,zhong2023clock,gromov2023modular,chughtai2023toy,stander2024grokking}, so that the delay is the time that it takes to build such a circuit. Accounts of the training dynamics see slow optimization that reshapes the network~\citep{liu2022understanding,kumar2024grokking,varma2024explaining,thilak2022slingshot,liu2022omnigrok}, and methods that shorten the delay treat it as wasted time~\citep{lee2024grokfast,walker2025grokalign,zhang2024hessian,xu2025groktransfer}. Each of these answers is drawn from a network that has already crossed the delay. For a general group, no one has computed before training which subspace the network is heading for.

This missing destination is what keeps the question open, because a hypothesis about the delay can be tested only if we know where the network is heading. We show that on group tasks, such as completing the multiplication table of a finite group, the destination can be computed from the task itself before training begins. It is a specific low-dimensional subspace for the embeddings of the network, which we call the \emph{target subspace}. We show that the symmetry of the task admits exactly one family of candidate subspaces, and that two requirements of the task select the target subspace within it, up to a tie that the theory itself identifies. Knowing the destination in advance lets us test our hypothesis that the delay is a search for the target subspace, and that the progress of this search sets when the network generalizes.

Because the destination is known before training, the hypothesis makes predictions that could fail (Figure~\ref{fig:anatomy}). First, if we restrict the network to the target subspace from the start, the delay must disappear. It does. With nothing else changed, the median grokking step falls from $64{,}050$ to $150$, before the network fits its training set, and the closer it starts to the target subspace, the shorter the delay. Second, if we ablate the target subspace, the network can no longer generalize. Its accuracy stops at a ceiling computed from the task in advance, and no trained network exceeds it.

Both experiments place the subspace by hand, so we also observe vanilla runs, to which nothing is done. They show the same search. During the delay, the embeddings move toward the target subspace, and the memorizing component of the logits fades as the generalizing component takes over. Moving the embeddings toward the target subspace during training brings generalization earlier, while an equally large move toward a random subspace barely does, and a hundredfold rise in gradient noise barely changes the delay.

\paragraph{Contributions.}
\begin{itemize}[leftmargin=1.2em,itemsep=1pt,topsep=2pt]
\item We propose that the grokking delay on group tasks is a search for the target subspace.
\item We compute the target subspace from the task before training, using the fact that the task's symmetry admits exactly one family of candidate subspaces (Theorem~\ref{thm:unique}).
\item We prove a parameter-free accuracy ceiling for any network whose inputs are restricted to an unfaithful set of irreducible representations (Theorem~\ref{thm:ceiling}).
\item We test the hypothesis by restricting the network to the target subspace, ablating it, observing vanilla training, and intervening during training.
\end{itemize}

\section{Related Work}
\label{sec:related}

\paragraph{Reverse-engineered destinations.} Networks that have grokked implement structured algorithms, such as Fourier circuits on modular addition~\citep{nanda2023progress,zhong2023clock,gromov2023modular} and representation circuits~\citep{chughtai2023toy} or a coset circuit~\citep{stander2024grokking} on general groups. \citet{wu2025proven} reconcile these descriptions, margin analyses derive them as the solution of maximum margin~\citep{morwani2024feature,shutman2025words}, and reading hidden units in the basis of the task's group reveals the structure earlier in training~\citep{sivasankar2026fsd}. Apart from the margin analyses, these works read the destination off a network after it has arrived. Theory for two-layer networks shows that each neuron converges to one irreducible representation~\citep{he2026spectral} and that irreducible representations are learned one at a time, in an order computed from the input encoding~\citep{seqgroup2026}. These results describe which representations neurons settle on and in what order, rather than the subspace that the task requires. \citet{mallinar2025kernel} watch the destination arrive in a non-neural model trained on modular arithmetic, where block-circulant features that implement the Fourier algorithm emerge gradually while the usual measures of performance stay flat. Their features are found after training, on an abelian task, whereas we compute the destination before training on non-abelian groups, bound the accuracy of every network denied it, and intervene on the approach.

\paragraph{Explanations of the delay.} A second line of work explains why the delay is long. The proposed causes are a phase transition~\citep{liu2022understanding}, weight norms~\citep{liu2022omnigrok,khanh2026firstpassage,khanh2026weightnorm}, a transition from lazy to rich training~\citep{kumar2024grokking,mohamadi2024whygrok}, a competition between circuits~\citep{varma2024explaining}, an optimizer instability~\citep{thilak2022slingshot}, and the acquisition of task symmetries~\citep{hwang2026symmetry}. Another analysis separates a slowly learned representation from a fast readout~\citep{chou2026twospeeds}. Each account identifies a cause of the delay, but none computes the destination before training. Feature-learning theory bounds the loss of a gradient-trained two-layer network by that of the best network built on the features that its gradients expose~\citep{shi2023provable}. Our ceiling is a result of the opposite kind. It bounds the accuracy of every network once its inputs are restricted, and so it separates slow learning, which more training cures, from lost information, which no training recovers. Outside grokking, delays are known to depend on the overlap between the initialization and a low-dimensional target that the experimenter plants~\citep{saxe2014exact,atanasov2022silent,saad1995online,benarous2021online,abbe2023sgd,barak2022hidden}.

\paragraph{Shortening the delay.} Some methods change the training procedure~\citep{lee2024grokfast,zhang2024hessian,walker2025grokalign,tveit2025muon}. Others place structure into the network, taken from a run that has already grokked~\citep{minegishi2023grokking,xu2025groktransfer}, or, in concurrent work, built from labels~\citep{howe2026priors}. Architectures that build symmetry in by design use the same operator as we do~\citep{cohen2016group,kondor2018generalization,marchetti2024harmonics}, but they commit to the symmetry in advance, whereas we ask where unconstrained training goes.

\section{Preliminaries}
\label{sec:prelim}

\subsection{Task and Data}
\label{sec:task}

Our setup follows \citet{stander2024grokking}, who reverse-engineer networks that grok the multiplication of permutation groups. A finite group $G$ is a set of $|G|$ elements with an associative composition rule and inverses. The multiplication table records the rule in full as the $|G|\times|G|$ grid with entry $ab$ at position $(a,b)$. Our running example is $S_5$, the $120$ permutations of five objects under composition. The task is to complete the multiplication table. The network receives a pair $(a,b)$ and must predict $f(a,b)=ab$, which is classification with $|G|$ classes. Three more tasks are conjugation $f(a,b)=aba^{-1}$, the commutator $f(a,b)=a^{-1}b^{-1}ab$, and the squared conjugation $f(a,b)=(aba^{-1})^{2}=ab^{2}a^{-1}$, each a short expression in $a$ and $b$.

A random $40\%$ of the $|G|^2$ pairs forms the training set, and the rest form the test set. We call this partition the \emph{split}. On $S_5$, it leaves $8{,}640$ test pairs. The test set contains no new elements, only new combinations of them, so the network can generalize only by learning the composition rule. Each input is a one-hot group element, so the group acts on the inputs by permuting them, and our construction uses only this action and the multiplication table.

\subsection{Model and Measurements}
\label{sec:model}

The base model is a one-hidden-layer network with separate embeddings for the two arguments,
\begin{equation}
z(a,b)=U^\top\mathrm{ReLU}\bigl(W_l\,E_l\,e_a+W_r\,E_r\,e_b+b_h\bigr),
\label{eq:net}
\end{equation}
where $e_a$ and $e_b$ are the one-hot vectors of $a$ and $b$, $b_h$ is the bias of the hidden layer, and $z(a,b)$ holds one logit per class. The embeddings $E_l$ and $E_r$ are $d\times|G|$ matrices whose columns are the vectors of the group elements, so each of their rows is a vector in $\R^{|G|}$, the space of the one-hot inputs. We call $\R^{|G|}$ the \emph{input space}, and every subspace in this paper lies in it. Restricting the embeddings to a subspace with orthogonal projector $P$ replaces each embedding matrix $E\in\{E_l,E_r\}$ by $EP$, which is the same as projecting each one-hot input onto the subspace before it is embedded. Unless stated otherwise, restricting the network also replaces the output weights $U$ by $UP$, so that the logits lie in the same subspace. We train with full-batch AdamW~\citep{kingma2015adam,loshchilov2019decoupled} and weight decay, and some experiments use transformers or other optimizers instead.

We track two events in each run. The \emph{memorization step} is the first step at which training accuracy reaches $1.0$, and the \emph{grokking step} $\tau$ is the first step at which test accuracy reaches $0.95$. The \emph{delay} is the gap between the two. In Figure~\ref{fig:delay}, every seed memorizes by step $400$, at a median of $390$, but the median grokking step is $64{,}050$, and individual seeds range from $27{,}600$ to $160{,}500$. We call runs that train without any intervention \emph{vanilla}.

\section{Method}
\label{sec:method}
\label{sec:construction}

\begin{hypothesis}
On group tasks, the grokking delay is a search for the target subspace, and the progress of this search sets when the network generalizes.
\end{hypothesis}

Testing the hypothesis requires the target subspace before training. We first construct it from the task alone and show what the task's symmetry fixes. We then derive two predictions that could fail, a law for the delay and a ceiling for the accuracy, each with a quantity fixed in advance.

\subsection{Projectors from the Group}
\label{sec:idempotent}

The target subspace is built from projectors that the group itself provides. We need three notions from representation theory. A \emph{representation} of $G$ assigns a matrix to each element so that composing elements corresponds to multiplying their matrices. Every representation of a finite group decomposes into irreducible representations, or \emph{irreps}. Irreps play the role that frequencies play in Fourier analysis. Just as a periodic signal splits into frequencies, a vector in the input space splits into parts that belong to different irreps. Each irrep $\rho_i$ has a dimension $d_i$ and a \emph{character} $\chi_i$, which lists for every element the trace of its matrix, the sum of the diagonal entries. A character is a list of $|G|$ numbers that can be computed from the multiplication table alone.

A single classical formula~\citep{serre1977linear,fultonharris1991} turns characters into projectors that extract these parts. Let $L_g$ be the permutation matrix that sends the basis vector of each element $x$ to that of $gx$. These matrices form the \emph{regular representation} of $G$. The projector of an irrep $\rho_i$, and of a set $S$ of irreps, is
\begin{equation}
P_i=\frac{d_i}{|G|}\sum_{g\in G}\overline{\chi_i(g)}\,L_g,
\qquad P_S=\sum_{i\in S}P_i .
\label{eq:idempotent}
\end{equation}
The matrix $P_S$ is the orthogonal projector onto the part of the input space that belongs to the irreps in $S$. Representation theory calls this part the isotypic component of $S$, and $P_i$ the central idempotent of $\rho_i$. Each irrep contributes $d_i^2$ dimensions. Because everything in Eq.~\eqref{eq:idempotent} comes from the multiplication table, $P_S$ is available before training. On the cyclic group $\Z_n$, where the task is modular addition, $P_S$ reduces to the cosine and sine of one Fourier frequency, so the familiar Fourier answer on modular addition is the abelian special case of our construction.

\subsection{The Target Subspace}
\label{sec:right}
\label{sec:writing}

Equation~\eqref{eq:idempotent} gives one projector for every set of irreps. Two requirements determine which sets give a target subspace.

The first requirement is \emph{faithfulness}. An irrep that sends an element other than the identity to the identity matrix cannot distinguish this element from the identity. The elements that every irrep in $S$ treats in this way form the \emph{kernel} of $S$,
\begin{equation}
K_S=\{\,g\in G:\ \rho_i(g)=I\ \text{for every } i\in S\,\},
\label{eq:kernel}
\end{equation}
and $S$ is \emph{faithful} if $K_S$ contains only the identity. An unfaithful set is blind to part of the group, and $|K_S|$ measures how blind it is. On $S_5$, the sign irrep sends each permutation to its parity, $+1$ or $-1$, so its kernel is the alternating group $A_5$ of the $60$ even permutations. A network that sees only the sign irrep knows the parity of each input and nothing else. We meet this irrep twice more, once as the irrep that caps accuracy over all input pairs at $1/60$ and once as the irrep that vanilla networks hold through most of the delay.

The second requirement is \emph{completeness}, which means that the embedding must contain the whole subspace of each chosen irrep, not just part of it. The reason is the margin. The \emph{max-margin readout} maximizes the worst-case gap between the correct class and the best wrong class at a fixed scale, and it scores each class $c$ by the characters of the chosen irreps~\citep{morwani2024feature},
\begin{equation}
s(c)=\sum_{i\in S}\kappa_i\,\chi_i\!\left(c^{-1}ab\right),\qquad\kappa_i>0,
\label{eq:readout}
\end{equation}
where complex-conjugate irreps share one weight so that the score is real, and each term takes its largest value, $\chi_i(e)=d_i$, at the correct class $c=ab$. Each term expands into products of the matrix entries of the irrep at the two inputs, so an embedding that keeps part of the irrep loses every product with a missing entry, and the term at the correct class falls from $d_i$ to $k$ when $k$ of the $d_i$ indices remain. A partial subspace still distinguishes every element, and a network restricted to one can even cross the grokking threshold, yet the max-margin solution is out of its reach, and the networks that we restricted to partial subspaces stalled short of full accuracy.

\begin{definition}[Target subspace]
\label{def:right}
A \emph{target subspace} of a group task is the complete subspace of a faithful set of irreps, that is, the image of $P_S$ for a faithful set $S$. A \emph{minimal target subspace} uses a faithful set of smallest total dimension.
\end{definition}

\subsection{Symmetry Determines the Candidates}
\label{sec:unique}

Definition~\ref{def:right} selects among the projectors $P_S$, but the input space contains a continuum of other subspaces, and the symmetry of the task determines which of them can serve. Multiplication is symmetric on both sides. Relabeling the inputs as $(a,b)\mapsto(ga,bh)$ relabels the answer as $ab\mapsto g\,ab\,h$. A relabeling changes the names of the elements but not the task, so a subspace that the task itself singles out must respect it on both sides. In the other three tasks, conjugating both arguments by the same element conjugates the answer, and we use the same family of subspaces for them. With $e_x$ the basis vector of element $x$, the left and right translations of the input space are the permutation matrices $L_g$ and $R_h$ with $L_ge_x=e_{gx}$ and $R_he_x=e_{xh}$. A projector $P$ respects the symmetry of the task if, for all $g,h\in G$,
\begin{equation}
L_gP=PL_g
\quad\text{and}\quad
R_hP=PR_h .
\label{eq:commute}
\end{equation}
One side would not suffice, since a projector commuting only with the translations of one side can still split an irrep.

\begin{theorem}[Symmetry determines the family]
\label{thm:unique}
A projector $P$ on the input space satisfies Eq.~\eqref{eq:commute} if and only if $P=P_S$ for a set $S$ of irreps, with complex irreps in conjugate pairs.
\end{theorem}

\begin{proof}[Proof sketch]
Commuting with every right translation makes the projector a left convolution by a vector. Commuting with every left translation makes it central in the group algebra. Idempotence forces its coefficients into $\{0,1\}$.
\end{proof}

Theorem~\ref{thm:unique} states exactly what the symmetry of the task fixes. No projector outside the family $\{P_S\}$ satisfies Eq.~\eqref{eq:commute}. Within the family, faithfulness and completeness select the target subspaces, and minimality selects among them, so the destination is unique up to a tie that the theory itself identifies. On $S_5$, exactly two minimal candidates remain, the two \emph{tied irreps}. Each spans a minimal target subspace of $16$ of the $120$ input dimensions, and the choice between them is the only freedom that our account leaves open.

Our contribution is not the classical argument behind the theorem, but the condition under which it applies and the line that it draws. The theorem is about the task, and every result that follows is about the optimizer. Nothing obliges gradient descent, which is never equivariant during training, to end in the family that the symmetry permits. Whether it does is the empirical question that makes the hypothesis testable, and it does. Vanilla runs end with a median of $0.905$ of their embedding in a minimal target subspace.

\subsection{Prediction 1: Restricting to the Target Subspace}
\label{sec:pred1}
\label{sec:predictions}

If the delay is the time that a search takes, a network that starts at the destination has nothing left to search for. We restrict the network to the target subspace at every forward pass, so that its embeddings cannot leave it, and the first prediction is that the delay then disappears.

The same reasoning applies by degrees. The \emph{alignment} of both embeddings with a subspace whose projector is $P$ is the fraction of their squared Frobenius norm in the subspace,
\begin{equation}
\alpha=\frac{\|E_lP\|_F^2+\|E_rP\|_F^2}{\|E_l\|_F^2+\|E_r\|_F^2},
\label{eq:alignment}
\end{equation}
so that an alignment of $1$ means that both embeddings lie entirely in the subspace. We write $\alpha_0$ for the alignment of the initial embeddings with the minimal target subspace. A random initialization on $S_5$ has an alignment of about $0.133$, because this subspace occupies $16$ of the $120$ dimensions, and we call this value the \emph{chance alignment}. A vanilla network thus starts with most of its embedding outside the target subspace, and a network that starts closer must have a shorter delay, in a graded and lawful way.

To fix in advance what \emph{lawful} means, we chose five candidate forms for the grokking step $\tau$ as a function of $\alpha_0$ and committed to the form that fits best. The form that we expected is
\begin{equation}
\tau(\alpha_0)=t_0+C\,\alpha_0^{-p},
\label{eq:alignlaw}
\end{equation}
a power law with a floor, where $t_0$, $C$ and $p$ are fitted constants and $t_0$ is an offset, not a measured time. A first-passage argument gives this form under two assumptions. A head start compounds, so the alignment grows at a rate proportional to $\alpha^{1+p}$, and grokking arrives a fixed time after the alignment crosses a threshold. Because the argument assumes this growth rather than deriving it from the optimizer, Eq.~\eqref{eq:alignlaw} describes how $\tau$ depends on $\alpha_0$, not the path of the alignment. The four alternatives were a pure power law and hyperbolic, exponential and logarithmic forms.

\subsection{Prediction 2: Ablating the Target Subspace}
\label{sec:pred2}

If the network can succeed only through the target subspace, a network without it cannot generalize, and how much of the task survives can be computed in advance. We \emph{ablate} the target subspace by restricting the embeddings to an unfaithful set $S$ instead. The restriction collapses information at the input, because two elements $a$ and $ak$ with $k\in K_S$ receive identical embeddings. These elements merge into cosets of $K_S$, groups of elements that the network can no longer tell apart, and every later layer sees only the pair of cosets to which an input belongs. Any network can then at best predict, for each pair of cosets, its most common answer, a count available from the multiplication table before training.

\begin{theorem}[Accuracy ceiling]
\label{thm:ceiling}
Let the inputs of any of the four tasks enter the network only through embeddings restricted to an unfaithful set $S$. Then the accuracy over all input pairs of any network, of any depth and trained for any time, is at most the fraction of input pairs whose answer is the most common answer within their pair of cosets.
\end{theorem}

\begin{corollary}[A single occurrence]
\label{cor:regular}
If one of the two arguments appears exactly once in the task, as in multiplication and conjugation, the ceiling of Theorem~\ref{thm:ceiling} equals $1/|K_S|$.
\end{corollary}

On $S_5$, the sign irrep shows only parity, so the $120$ elements merge into the even and the odd coset. Each of the four pairs of cosets holds $3{,}600$ input pairs whose products are the $60$ elements of one coset, each occurring $60$ times, so the best bet is right on one pair in sixty, and the ceiling is $1/60$.

The ceiling has no free parameters, and the larger the kernel, the lower the ceiling. It holds for any architecture after the embedding, since the condition is on the input pathway alone, and the same count over the test pairs bounds the test accuracy. It is not always small. A kernel in the center of $G$ leaves the commutator unchanged and the ceiling at $1$.

\section{Experiments}
\label{sec:experiments}

We test four consequences of the hypothesis, one per subsection. Restricting the network to the target subspace should remove the delay, ablating it should cap accuracy at the ceiling of Theorem~\ref{thm:ceiling}, vanilla training should move toward it during the delay, and moving the embeddings toward it should shorten the delay. Unless noted, the task is $S_5$ multiplication and numbers are medians over seeds.

\subsection{Restricting to the Target Subspace}
\label{sec:supply}

Prediction~1 is the most direct test of the hypothesis, so we tested it with nothing else changed. We restricted the network to the minimal target subspace of $S_5$, embeddings and output weights alike. The delay disappeared on all ten seeds (Figure~\ref{fig:anatomy}, left). The restricted network reached $0.95$ test accuracy at a median of $150$ steps, against $64{,}050$ for the twenty vanilla runs, and it did so before it had fit its training set at $220$ steps, so it never needed to memorize first. Neither weight decay nor the projection explains this. Without weight decay, the restricted network grokked at step $200$ on all five seeds, while vanilla training did not grok in $100$k steps. A soft penalty toward the target subspace, with no projection, grokked within $800$ steps at each of four strengths, and a network projected once at initialization grokked at $240$ to $270$ steps. What removes the delay is where the network starts, and not how it is put there.

\begin{figure}[t]
\centering
\includegraphics[width=0.995\columnwidth]{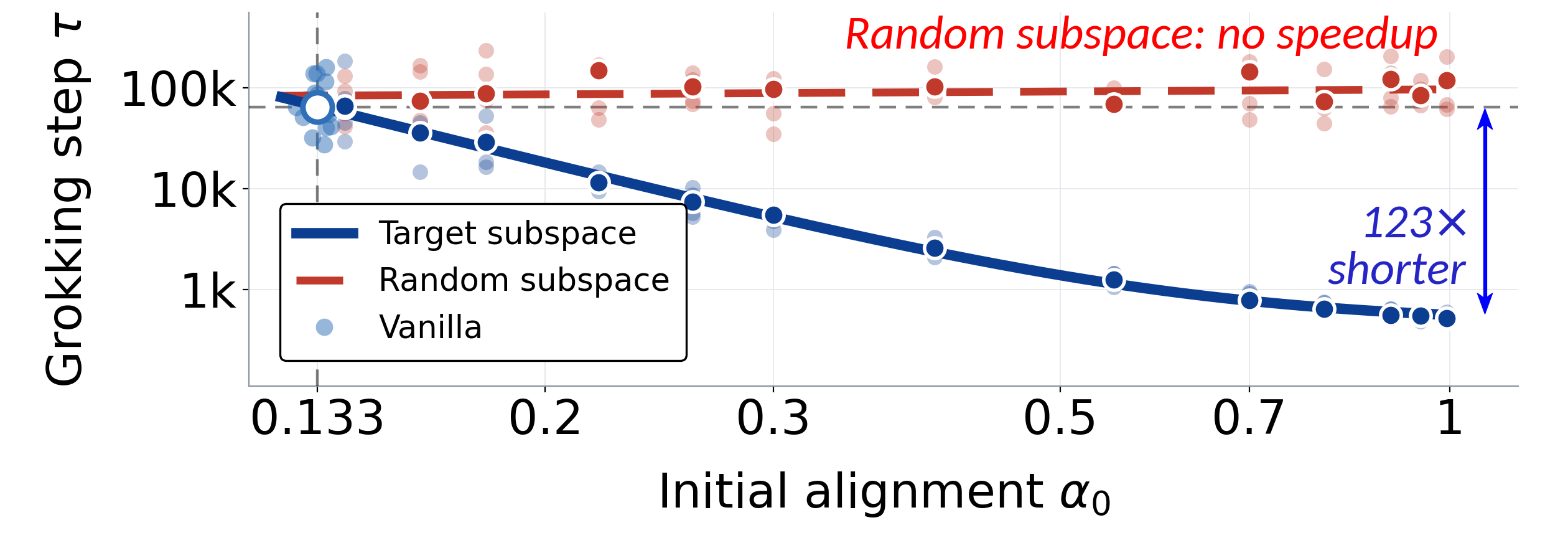}
\caption{\textbf{Starting closer shortens the delay.} The solid curve is Eq.~\eqref{eq:alignlaw}, and the open circle marks the vanilla median. Runs rotated toward a random subspace are plotted at their alignment with it.}
\label{fig:law}
\end{figure}

The graded form of the prediction is a stronger test, because Eq.~\eqref{eq:alignlaw} was fixed before the data. We rotated the initial embeddings toward the minimal target subspace to thirteen alignment levels $\alpha_0$ between $0.14$ and $0.995$, five seeds each, preserving all norms. The delay shrank by more than two orders of magnitude along the predicted curve (Figure~\ref{fig:law}). Among the five candidate forms, Eq.~\eqref{eq:alignlaw} fit the sixty-five rotated runs and the twenty vanilla runs best, by $\Delta\mathrm{AIC}=32.6$ over the nearest alternative, and still by $30.2$ when the alternatives were also given a floor. The strongest evidence is a prediction rather than a fit. No rotated run started at the chance alignment, so we fitted Eq.~\eqref{eq:alignlaw} on the rotated runs alone and asked where it placed vanilla training. It placed vanilla training at $63{,}158$ steps, against the observed $64{,}050$, an error of $1.4\%$, while the four rejected forms, refitted in the same way, missed by $29\%$ or more.

\begin{table}[t]
\centering\footnotesize
\caption{\textbf{Restricting the network to different subspaces of $S_5$.} Accuracies are median best values, and $\tau$ is the median grokking step of the runs that grok. A dash means no run grokked.}\label{tab:specificity}
\setlength{\tabcolsep}{4pt}
\begin{tabular*}{\columnwidth}{@{\extracolsep{\fill}}lrrrr@{}}
\toprule
 & & \multicolumn{2}{c}{Accuracy (\%)} & Grokking \\
\cmidrule(lr){3-4}
Subspace & Dim. & Train & Test & step $\tau$ \\
\midrule
Full space (vanilla) & $120$ & $100.0$ & $100.0$ & $64{,}050$ \\
Target subspace & $16$ & $100.0$ & $100.0$ & $150$ \\
\midrule
Random subspace & $16$ & $14.0$ & $0.8$ & --- \\
Principal directions & $16$ & $100.0$ & $96.9$ & $3{,}000$ \\
\midrule
Half of target subspace & $8$ & $35.7$ & $23.4$ & --- \\
Coset basis & $5$ & $1.7$ & $1.7$ & --- \\
Completed coset basis & $17$ & $100.0$ & $100.0$ & $150$ \\
\bottomrule
\end{tabular*}
\end{table}

These results would mean little if any concentrated subspace had the same effect, so we compared the target subspace with alternatives of the same dimension. Rotating toward a random subspace of the same dimension, at the same levels and seeds, concentrates exactly as much of the embedding in sixteen dimensions, yet the delay did not move (Figure~\ref{fig:law}). The slope of the log grokking step against the log initial alignment was $+0.07$, with an interval of $[-0.07,+0.18]$. Other restrictions gave the same verdict (Table~\ref{tab:specificity}). Of three $16$-dimensional subspaces, only the target subspace removed the delay. A random subspace, which lies outside the family $\{P_S\}$, kept the network from fitting its training set. The top sixteen principal directions of a converged network stalled at $0.97$, as completeness predicts for a subspace that holds only part of an irrep, until the missing directions were added. Half of the target subspace and the coset basis found by reverse engineering failed even to fit the training set, whereas the completed coset basis, which contains a minimal target subspace, grokked at $150$ steps like the target subspace. Freely trained embeddings of rank up to $48$ grokked less often than vanilla training in the same budget. The delay depends on which subspace holds the embedding, and not on how concentrated the embedding is.

The effect belongs to the task, and not to one setting. It recurred on twelve groups of order up to $720$ and on transformers of two to twelve layers. The data split is the cleanest test, because it changes the training set but not the task. Across five random splits, the vanilla ten-seed median ranged from $73$k to $85$k steps, while the restricted median stayed at $150$, so the split changes how long the search takes, but not where it ends. Under mini-batch AdamW and SGD with momentum, vanilla training memorized and generalized almost together, leaving no delay to shorten, yet the target subspace brought generalization $74$ and $22$ times sooner, and starting closer still helped. The delay itself is therefore a ratio of two speeds. In epochs, mini-batch training generalized about forty times sooner than full-batch training, but memorized about three times later.

Changing the training procedure does not come close. Six published accelerators, grid-searched under one protocol and budget, stayed one to three orders of magnitude behind wherever they grokked at all, the best at $4{,}250$ steps against $150$. Because the target subspace uses oracle information, we use it to diagnose the delay, not as an accelerator.

\subsection{Ablating the Target Subspace}
\label{sec:corrupt}

\begin{figure}[t]
\centering
\includegraphics[width=0.8\columnwidth]{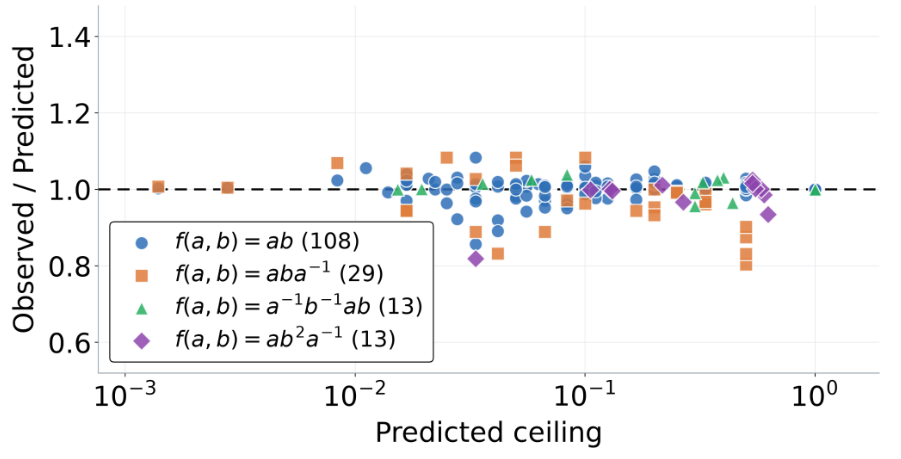}
\caption{\textbf{The plateaus follow the computed ceilings.} Plateau of each of the $163$ restricted configurations, divided by its ceiling computed over all input pairs. Points above one stay within the bound computed on their own test split.}
\label{fig:ceiling}
\end{figure}

The second prediction comes with a number, so it can fail by being wrong, and not only by being absent. We ablated the target subspace by restricting the network to an unfaithful set at every training step, and compared the \emph{plateau}, the accuracy at which the restricted network settles at the end of training, with the ceiling of Theorem~\ref{thm:ceiling} computed before training. The ablated network never grokked (Figure~\ref{fig:anatomy}, left). To test the ceiling far beyond one example, we ran $163$ restricted configurations over the four tasks and seventeen abelian and non-abelian groups, with ceilings spanning three orders of magnitude. The plateaus followed the computed ceilings across the whole range (Figure~\ref{fig:ceiling}). Over the $157$ configurations with a nontrivial kernel, they lay $3.0\%$ from the computed level on average and never more than a fifth away. The level is the $1/|K_S|$ of Corollary~\ref{cor:regular} on multiplication and conjugation and the count of Theorem~\ref{thm:ceiling} on the commutator and the squared conjugation. No plateau exceeded the bound on its own test split, which is the same count taken over the test pairs, so the plateaus that lie above the level computed over all pairs, $87$ of $157$ and by at most $8\%$, do not violate the theorem.

The sharpest evidence is that some plateaus are hit exactly, to the last test pair. Where the output weights are projected together with the embeddings, every class in a coset receives the same logit. Once the network predicts the correct output coset, no amount of training can separate the classes in it, so the number of test pairs that it gets right is fixed by the split and by the fixed rule that breaks ties in the argmax. On $S_5$, ablating every faithful irrep left the network $149$ of the $8{,}640$ test pairs, and ablating the sign irrep as well left it $75$, on all ten seeds with zero variance. We then changed the split, which changes the counts. On five random splits the pair of counts came out different every time, and every time the trained networks landed on them. Twenty-four of the twenty-six commutator and squared-conjugation configurations were also hit exactly. Their kernel covers the whole group, so the network can give only one answer, and in these configurations it gave one of the most frequent labels among the test pairs. A prediction that is met to the test pair, split after split, leaves no room for a fitted constant.

Theorem~\ref{thm:ceiling} constrains only what reaches the network, so nothing after the embedding should move the ceiling, and we changed the three things most likely to move it. A six-layer transformer reached the sign-irrep plateau on all five seeds, mini-batch AdamW reached the same plateau, and resampling $30\%$ of the training labels left the plateaus unchanged to four decimals, while the network restricted to the target subspace still grokked in each case. The failure therefore comes from the blindness of the unfaithful set, and not from the difficulty of the setting, because ablation removes information that no training recovers.

\subsection{Observing Vanilla Training}
\label{sec:observe}

The first two experiments placed the subspace by hand. If the hypothesis is right, the same search must be visible in vanilla runs, so we observed the twenty vanilla runs, using the projector only to measure. We center the logits $z_c(a,b)$ for each input pair and fit them by least squares, one term per irrep, in the form of the max-margin readout of Eq.~\eqref{eq:readout},
\begin{equation}
z_c(a,b)=\sum_i w_i\,\chi_i\!\left(c^{-1}ab\right)+r_c(a,b).
\label{eq:decomposition}
\end{equation}
We call the sum the \emph{generalizing component} and the residual $r$ the \emph{memorizing component}, because the argmax of $r$ over classes has an early accuracy close to the memorized training fraction. The right panel of Figure~\ref{fig:anatomy} plots this accuracy and the fraction of the logit variance that the generalizing component explains. The \emph{share} of an irrep is its fraction of the squared norm of the generalizing component.

\begin{figure}[t]
\centering
\includegraphics[width=\columnwidth]{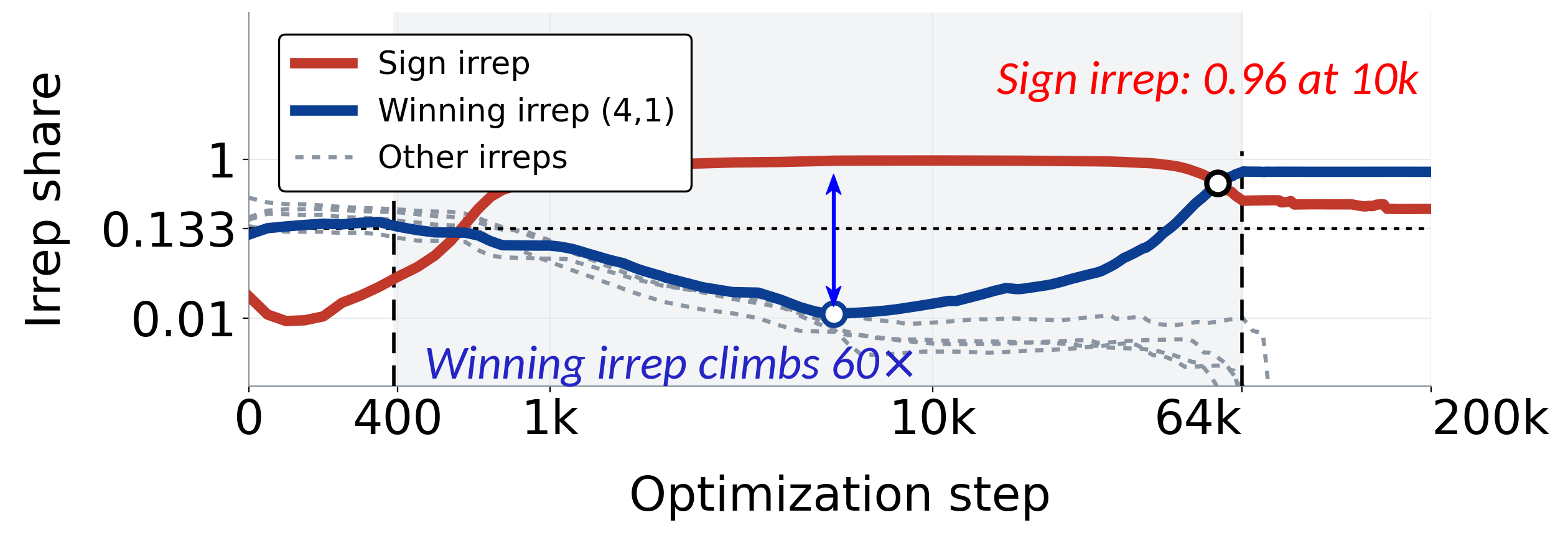}
\caption{\textbf{The delay is spent on the sign irrep.} Share of each irrep in the generalizing component, medians over the twenty vanilla runs. The dotted line, $16/120$, is the share that the $16$ dimensions of the winning irrep would receive by dimension alone.}
\label{fig:search}
\end{figure}

The delay is spent on an irrep that cannot do the task. Midway through the delay, almost all the generalizing component sat on the sign irrep, which held a share of $0.96$ at step $10$k (Figure~\ref{fig:search}) and by Corollary~\ref{cor:regular} cannot carry the task past an accuracy of $1/60$ over all input pairs. The memorizing component was no better. It peaked near the memorized training fraction, fell to $0.02$ while test accuracy was still below $3\%$, and reached zero only as test accuracy took off. Meanwhile, the tied irrep that training finally selects climbed about $60$ times between a floor in the middle of the delay and the grokking step. We call it the \emph{winning irrep}. It overtook the sign irrep only at the end of the delay, as the memorizing component vanished and test accuracy rose.

The network reaches the target subspace without any help. With the embeddings left fully free, the twenty converged embeddings placed a median of $0.905$ of their squared Frobenius norm in the minimal target subspace of the winning irrep, against the chance alignment of $0.133$. At the first measurement at or after each run's grokking step, the median was only $0.545$, so a run generalizes once it is close enough to the target subspace, and not once it has arrived. Eight of the twenty runs settled between $0.33$ and $0.57$, still $2.5$ to $4.3$ times the chance alignment, and generalized fully anyway, and the two least aligned runs were the two slowest.

The only freedom that the theory leaves is the tie between the two tied irreps, and training broke it the same way in all twenty runs, whichever tied irrep started with the larger alignment. The winning irrep is the irrep of the coset circuit identified by reverse engineering~\citep{stander2024grokking,wu2025proven}, so the destination that we compute before training is the circuit that others found after it. The outcome is a default, not a rule. An initial alignment of $0.15$ on the other tied irrep flipped it in three of ten runs, $0.165$ in four of five, and $0.18$ in all five. A nudge inside the family moves the destination, while no subspace outside it reached full accuracy (Table~\ref{tab:specificity}), which is exactly the latitude that Theorem~\ref{thm:unique} leaves open.

\subsection{Intervening During Training}
\label{sec:dynamics}

\begin{figure}[t]
\centering
\includegraphics[width=\columnwidth]{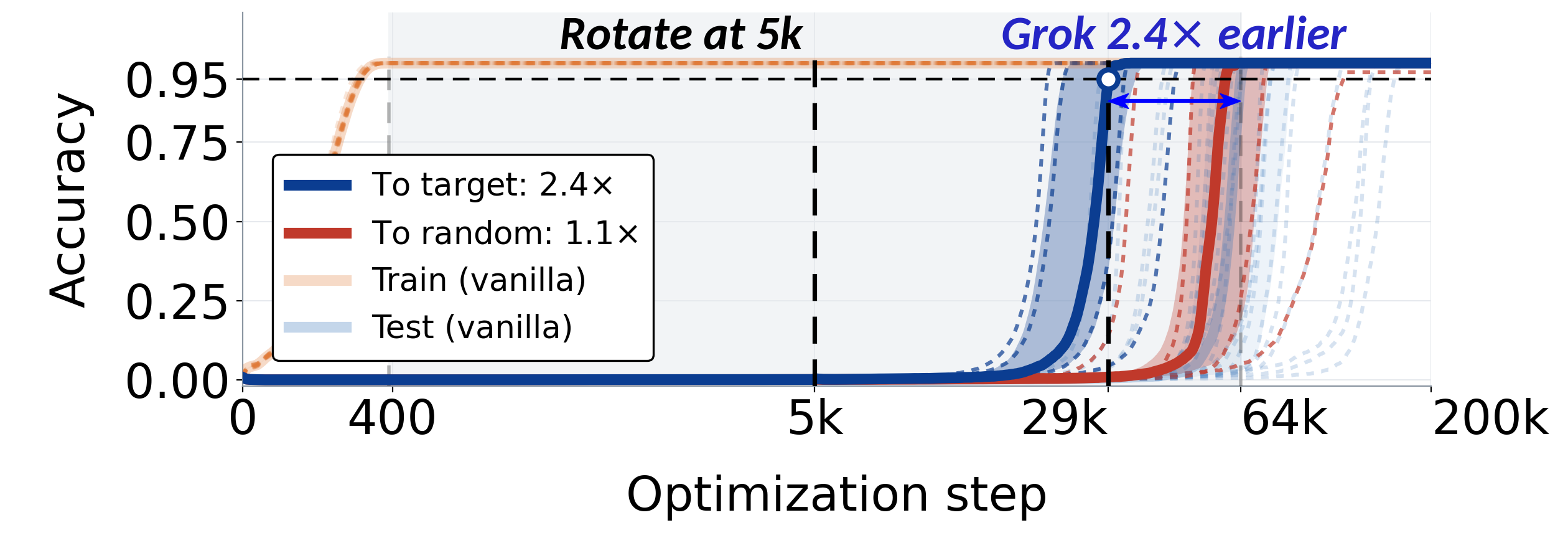}
\caption{\textbf{The alignment is a cause, not a symptom.} Test accuracy of twenty vanilla runs resumed at $5$k after rotating the embeddings by the same angle toward the target subspace or a random subspace of the same dimension. The legend gives paired median speedups.}
\label{fig:perturb}
\end{figure}

Vanilla runs approach the target subspace during the delay, but this correlation does not show that the approach sets the delay. The delay could instead be an escape from the memorizing solution, with the alignment merely a symptom of its progress. If so, changing the alignment should leave the delay unchanged. We therefore paused vanilla runs at step $5$k, rotated their embeddings toward the target subspace with nothing else changed, and resumed training. Resuming without rotating reproduced the parent run step for step, so any change comes from the rotation, not the restart.

Rotating to an alignment of $0.7$ shortened the delay on all five seeds, by $2.2\times$ in the paired median, and rotating to $0.4$ shortened it by $1.3\times$. The decisive control keeps the size of the move and changes only its direction. On twenty seeds, a rotation by exactly the same angle toward a random subspace of the same dimension shortened the delay by only $1.10\times$, against $2.40\times$ toward the target subspace (Figure~\ref{fig:perturb}). The rotation toward the target subspace was faster on every seed and stayed faster when the optimizer state was reset in every arm. Rotating to $0.7$ at step $30$k instead of $5$k cut the delay by $1.4\times$, so an earlier intervention buys more, as it should if the alignment measures how far the search has come. The alignment is a cause, not a symptom.

Noise does not set the delay either, so the delay is not a noise-driven escape. Full-batch training groks, and a hundredfold increase in the power of the gradient noise changed the delay by a factor of only $1.10$. Noise and weight decay change the speed of the search instead. Across the two sweeps, the grokking step of runs that started at the chance alignment ranged from $45{,}700$ to $137{,}600$, mostly through weight decay. In every setting, the slope of the log grokking step against the log initial alignment lay between $-2.4$ and $-2.6$, and no interval contained zero. Under the projector, the same two settings moved the grokking step by only $130$ steps. They lengthen the search, not what follows it.

\section{Discussion}
\label{sec:discussion}

Our results separate the delay from the destination. Under a different optimizer, the delay can vanish while the destination stays fixed, still setting when the network generalizes and capping accuracy at the same precomputed ceiling. On this evidence, accounts that locate grokking in optimization describe the speed of the search rather than its object.

Naming the destination also names what the delay is spent on. Most vanilla runs spend it with the generalizing component of their logits on the sign irrep, whose blindness we quantify before training. What looks like idle time from the outside is a network holding a representation that cannot do the task. Our observations agree with circuit accounts, which describe a competition between a memorizing and a generalizing solution, and what our account adds is the identity of the competitors, computed before training.

\paragraph{Limitations.} Our scope is that of a visible group action. Behind an ordinary encoder, the features lose the regular representation, and the construction no longer applies. On tasks without algebraic structure, the target subspace is undefined. Within this scope, three things remain open. First, ablating the target subspace tests the computed ceiling rather than the search, so the claim that the delay is a search rests on the measurements of the grokking step. Second, the mechanism behind Eq.~\eqref{eq:alignlaw} is not established. The equation relates $\tau$ to $\alpha_0$ without describing the path of the alignment, and its exponent ranges from $3.16$ to $4.18$ across fits and data splits, so what replicates is the form. Third, why one tied irrep wins on every seed of the split is unknown.

\section{Conclusion}
\label{sec:conclusion}

We computed the destination of training on group tasks before training and tested what a network does during the delay. Restricting the network to the target subspace removes the delay, ablating it caps accuracy at a ceiling computed in advance, vanilla networks move toward it long before they generalize, and moving them toward it brings generalization earlier. The delay is therefore not an intrinsic cost of generalization. Transfer to natural data is the largest open question. On every group task that we tested, grokking is a search for the target subspace, and the delay is the time that this search takes.

\newpage
\bibliography{references}

@article{power2022grokking,
  title={Grokking: Generalization beyond overfitting on small algorithmic datasets},
  author={Power, Alethea and Burda, Yuri and Edwards, Harri and Babuschkin, Igor and Misra, Vedant},
  journal={arXiv preprint arXiv:2201.02177},
  year={2022}
}

@inproceedings{liu2022understanding,
  title={Towards understanding grokking: An effective theory of representation learning},
  author={Liu, Ziming and Kitouni, Ouail and Nolte, Niklas and Michaud, Eric J. and Tegmark, Max and Williams, Mike},
  booktitle={Advances in Neural Information Processing Systems (NeurIPS)},
  year={2022}
}

@inproceedings{barak2022hidden,
  title={Hidden progress in deep learning: {SGD} learns parities near the computational limit},
  author={Barak, Boaz and Edelman, Benjamin L. and Goel, Surbhi and Kakade, Sham and Malach, Eran and Zhang, Cyril},
  booktitle={Advances in Neural Information Processing Systems (NeurIPS)},
  year={2022}
}

@inproceedings{liu2022omnigrok,
  title={Omnigrok: Grokking beyond algorithmic data},
  author={Liu, Ziming and Michaud, Eric J. and Tegmark, Max},
  booktitle={International Conference on Learning Representations (ICLR)},
  year={2023}
}

@inproceedings{nanda2023progress,
  title={Progress measures for grokking via mechanistic interpretability},
  author={Nanda, Neel and Chan, Lawrence and Lieberum, Tom and Smith, Jess and Steinhardt, Jacob},
  booktitle={International Conference on Learning Representations (ICLR)},
  year={2023}
}

@article{varma2024explaining,
  title={Explaining grokking through circuit efficiency},
  author={Varma, Vikrant and Shah, Rohin and Kenton, Zachary and Kram{\'a}r, J{\'a}nos and Kumar, Ramana},
  journal={arXiv preprint arXiv:2309.02390},
  year={2023}
}

@inproceedings{kumar2024grokking,
  title={Grokking as the transition from lazy to rich training dynamics},
  author={Kumar, Tanishq and Bordelon, Blake and Gershman, Samuel J. and Pehlevan, Cengiz},
  booktitle={International Conference on Learning Representations (ICLR)},
  year={2024}
}

@inproceedings{mohamadi2024whygrok,
  title={Why do you grok? {A} theoretical analysis of grokking modular addition},
  author={Mohamadi, Mohamad Amin and Li, Zhiyuan and Wu, Lei and Sutherland, Danica J.},
  booktitle={International Conference on Machine Learning (ICML)},
  year={2024}
}

@article{gromov2023modular,
  title={Grokking modular arithmetic},
  author={Gromov, Andrey},
  journal={arXiv preprint arXiv:2301.02679},
  year={2023}
}

@article{thilak2022slingshot,
  title={The Slingshot Mechanism: An Empirical Study of Adaptive Optimizers and the Grokking Phenomenon},
  author={Thilak, Vimal and Littwin, Etai and Zhai, Shuangfei and Saremi, Omid and Paiss, Roni and Susskind, Joshua},
  journal={arXiv preprint arXiv:2206.04817},
  year={2022}
}

@inproceedings{morwani2024feature,
  title={Feature emergence via margin maximization: Case studies in algebraic tasks},
  author={Morwani, Depen and Edelman, Benjamin L. and Oncescu, Costin-Andrei and Zhao, Rosie and Kakade, Sham},
  booktitle={International Conference on Learning Representations (ICLR)},
  year={2024}
}

@inproceedings{mallinar2025kernel,
  title={Emergence in non-neural models: Grokking modular arithmetic via average gradient outer product},
  author={Mallinar, Neil and Beaglehole, Daniel and Zhu, Libin and Radhakrishnan, Adityanarayanan and Pandit, Parthe and Belkin, Mikhail},
  booktitle={International Conference on Machine Learning (ICML)},
  year={2025}
}

@inproceedings{chughtai2023toy,
  title={A toy model of universality: Reverse engineering how networks learn group operations},
  author={Chughtai, Bilal and Chan, Lawrence and Nanda, Neel},
  booktitle={International Conference on Machine Learning (ICML)},
  year={2023}
}

@inproceedings{stander2024grokking,
  title={Grokking group multiplication with cosets},
  author={Stander, Dashiell and Yu, Qinan and Fan, Honglu and Biderman, Stella},
  booktitle={International Conference on Machine Learning (ICML)},
  year={2024}
}

@inproceedings{wu2025proven,
  title={Towards a Unified and Verified Understanding of Group-Operation Networks},
  author={Wu, Wilson and Jaburi, Louis and Drori, Jacob and Gross, Jason},
  booktitle={International Conference on Learning Representations (ICLR)},
  year={2025},
  note={arXiv:2410.07476}
}

@inproceedings{cohen2016group,
  title={Group equivariant convolutional networks},
  author={Cohen, Taco and Welling, Max},
  booktitle={International Conference on Machine Learning (ICML)},
  pages={2990--2999},
  year={2016}
}

@inproceedings{kondor2018generalization,
  title={On the generalization of equivariance and convolution in neural networks to the action of compact groups},
  author={Kondor, Risi and Trivedi, Shubhendu},
  booktitle={International Conference on Machine Learning (ICML)},
  pages={2747--2755},
  year={2018}
}

@book{serre1977linear,
  title={Linear representations of finite groups},
  author={Serre, Jean-Pierre},
  volume={42},
  year={1977},
  publisher={Springer-Verlag},
  address={New York}
}

@book{fultonharris1991,
  title={Representation theory: A first course},
  author={Fulton, William and Harris, Joe},
  volume={129},
  year={1991},
  publisher={Springer},
  series={Graduate Texts in Mathematics}
}

@article{lee2024grokfast,
  title={Grokfast: Accelerated grokking by amplifying slow gradients},
  author={Lee, Jaerin and Kang, Bong Gyun and Kim, Kihoon and Lee, Kyoung Mu},
  journal={arXiv preprint arXiv:2405.20233},
  year={2024}
}

@inproceedings{xu2025groktransfer,
  title={Let Me Grok for You: Accelerating Grokking via Embedding Transfer from a Weaker Model},
  author={Xu, Zhiwei and Ni, Zhiyu and Wang, Yixin and Hu, Wei},
  booktitle={International Conference on Learning Representations (ICLR)},
  year={2025},
  note={arXiv:2504.13292}
}

@article{walker2025grokalign,
  title={GrokAlign: Geometric Characterisation and Acceleration of Grokking},
  author={Walker, Thomas and Humayun, Ahmed Imtiaz and Balestriero, Randall and Baraniuk, Richard},
  journal={arXiv preprint arXiv:2506.12284},
  year={2025}
}

@article{hwang2026symmetry,
  title={Intrinsic Task Symmetry Drives Generalization in Algorithmic Tasks},
  author={Hwang, Hyeonbin and Park, Yeachan},
  journal={arXiv preprint arXiv:2603.01968},
  year={2026}
}

@inproceedings{seqgroup2026,
  title={Sequential Group Composition: A Window into the Mechanics of Deep Learning},
  author={Marchetti, Giovanni Luca and Kunin, Daniel and Myers, Adele and Acosta, Francisco and Miolane, Nina},
  booktitle={International Conference on Machine Learning (ICML)},
  year={2026},
  note={arXiv:2602.03655}
}

@inproceedings{kingma2015adam,
  title={{Adam}: A method for stochastic optimization},
  author={Kingma, Diederik P. and Ba, Jimmy},
  booktitle={International Conference on Learning Representations (ICLR)},
  year={2015}
}

@inproceedings{loshchilov2019decoupled,
  title={Decoupled weight decay regularization},
  author={Loshchilov, Ilya and Hutter, Frank},
  booktitle={International Conference on Learning Representations (ICLR)},
  year={2019}
}

@inproceedings{foret2021sharpness,
  title={Sharpness-Aware Minimization for Efficiently Improving Generalization},
  author={Foret, Pierre and Kleiner, Ariel and Mobahi, Hossein and Neyshabur, Behnam},
  booktitle={International Conference on Learning Representations (ICLR)},
  year={2021}
}

@inproceedings{zhang2024hessian,
  title={A Hessian View of Grokking in Mathematical Reasoning},
  author={Zhang, Zhenshuo and Liu, Jerry W. and R{\'e}, Christopher and Zhang, Hongyang R.},
  booktitle={The 4th Workshop on Mathematical Reasoning and AI at NeurIPS 2024 (MATH-AI)},
  year={2024}
}

@article{tveit2025muon,
  title={Muon Optimizer Accelerates Grokking},
  author={Tveit, Amund and Remseth, Bj{\o}rn and Skogvold, Arve},
  journal={arXiv preprint arXiv:2504.16041},
  year={2025}
}

@article{shutman2025words,
  title={Learning words in groups: fusion algebras, tensor ranks and grokking},
  author={Shutman, Maor and Louidor, Oren and Tessler, Ran},
  journal={arXiv preprint arXiv:2509.06931},
  year={2025}
}

@inproceedings{zhong2023clock,
  title={The Clock and the Pizza: Two Stories in Mechanistic Explanation of Neural Networks},
  author={Zhong, Ziqian and Liu, Ziming and Tegmark, Max and Andreas, Jacob},
  booktitle={Advances in Neural Information Processing Systems (NeurIPS)},
  year={2023}
}

@inproceedings{marchetti2024harmonics,
  title     = {Harmonics of Learning: Universal {F}ourier Features Emerge in Invariant Networks},
  author    = {Marchetti, Giovanni Luca and Hillar, Christopher J and Kragic, Danica and Sanborn, Sophia},
  booktitle = {Proceedings of the Thirty Seventh Conference on Learning Theory},
  series    = {Proceedings of Machine Learning Research},
  volume    = {247},
  pages     = {3775--3797},
  publisher = {PMLR},
  year      = {2024}
}

@inproceedings{saxe2014exact,
  title={Exact solutions to the nonlinear dynamics of learning in deep linear neural networks},
  author={Saxe, Andrew M. and McClelland, James L. and Ganguli, Surya},
  booktitle={International Conference on Learning Representations (ICLR)},
  year={2014}
}

@inproceedings{atanasov2022silent,
  title={Neural networks as kernel learners: The silent alignment effect},
  author={Atanasov, Alexander and Bordelon, Blake and Pehlevan, Cengiz},
  booktitle={International Conference on Learning Representations (ICLR)},
  year={2022}
}

@article{saad1995online,
  title={On-line learning in soft committee machines},
  author={Saad, David and Solla, Sara A.},
  journal={Physical Review E},
  volume={52},
  number={4},
  pages={4225--4243},
  year={1995}
}

@article{benarous2021online,
  title={Online stochastic gradient descent on non-convex losses from high-dimensional inference},
  author={Ben Arous, G{\'e}rard and Gheissari, Reza and Jagannath, Aukosh},
  journal={Journal of Machine Learning Research},
  volume={22},
  number={106},
  pages={1--51},
  year={2021}
}

@inproceedings{abbe2023sgd,
  title={{SGD} learning on neural networks: leap complexity and saddle-to-saddle dynamics},
  author={Abbe, Emmanuel and Boix-Adser{\`a}, Enric and Misiakiewicz, Theodor},
  booktitle={Conference on Learning Theory (COLT)},
  pages={2552--2623},
  year={2023}
}

@article{minegishi2023grokking,
  title={Bridging Lottery Ticket and Grokking: Understanding Grokking from Inner Structure of Networks},
  author={Minegishi, Gouki and Iwasawa, Yusuke and Matsuo, Yutaka},
  journal={Transactions on Machine Learning Research},
  year={2025}
}

@article{howe2026priors,
  title={Structure-Specific Representational Priors Causally Control the Grokking Delay},
  author={Howe, Gunner Levi},
  journal={arXiv preprint arXiv:2607.04333},
  year={2026},
  note={Concurrent preprint}
}

@article{sivasankar2026fsd,
  title  = {Learned Structure Shows Up Earlier in the Task's Own Basis: A Controlled Look at a Synchronisation Precursor in Grokking},
  author = {Sivasankar, Achyuthan},
  journal= {arXiv preprint arXiv:2606.12966},
  year   = {2026},
  note   = {v3; concurrent preprint}
}

@inproceedings{shi2023provable,
  title={Provable Guarantees for Neural Networks via Gradient Feature Learning},
  author={Shi, Zhenmei and Wei, Junyi and Liang, Yingyu},
  booktitle={Advances in Neural Information Processing Systems (NeurIPS)},
  year={2023}
}

@article{he2026spectral,
  title={Neural Networks Provably Learn Spectral Representations for Group Composition},
  author={He, Jianliang and Wang, Leda and Zhang, Fengzhuo and Chen, Siyu and Yang, Zhuoran},
  journal={arXiv preprint arXiv:2606.02993},
  year={2026}
}

@article{khanh2026firstpassage,
  title={First-Passage Prediction of Grokking Delay: A Calibrated Law under {AdamW} with Causal Validation},
  author={Khanh, Truong Xuan and Hoa, Truong Quynh and Trung, Luu Duc and Duc, Phan Thanh},
  journal={arXiv preprint arXiv:2605.18845},
  year={2026}
}

@article{khanh2026weightnorm,
  title={The Weight Norm Sets the Grokking Timescale: A Causal Delay Law},
  author={Khanh, Truong Xuan and Viet, Doan Hoang and Trung, Luu Duc and Duc, Phan Thanh},
  journal={arXiv preprint arXiv:2606.13753},
  year={2026}
}

@article{chou2026twospeeds,
  title={Two Speeds of Learning: A Representation-Readout Decomposition of Grokking and Double Descent},
  author={Chou, Chi-Ning and Uzdelewicz, Oscar and Chiu, Neng-Chun and Yang, Yao-Yuan and Chung, SueYeon},
  journal={arXiv preprint arXiv:2605.27078},
  year={2026}
}
\bibliographystyle{icml2027}

\appendix
\let\addcontentsline\origaddcontentsline 
\setcounter{tocdepth}{2}
\twocolumn[{\centering\Large\textbf{Grokking as a Subspace Search: Computing the Destination Before Training}\\\vspace{0.5em}Supplementary Material\\\vspace{1.2em}
  \begin{minipage}{0.92\textwidth}\normalsize\raggedright
  {\large\textbf{Table of Contents}}\\[-0.4em]\rule{\linewidth}{0.6pt}
  \makeatletter\renewcommand{\contentsname}{}\makeatother
  \vspace{-2.6em}\tableofcontents\vspace{-0.2em}
  \rule{\linewidth}{0.6pt}
  \end{minipage}\vspace{1.5em}}]

\section{Completing the Construction}
\label{app:recipe}

This section carries out the construction of Section~\ref{sec:construction} in full. It gives the projector entry by entry, the kernel that it defines, the abelian case that recovers the familiar Fourier answer, and the quantitative form of the completeness requirement.

\paragraph{Setup.} Let $G$ be a finite group of order $n=|G|$ and let $L$ be its regular representation on $\R^{|G|}$, with basis $\{e_x\}_{x\in G}$ and $L_ge_x=e_{gx}$ as in Eq.~\eqref{eq:idempotent}. To decompose $L$ we work over $\C$, extending the input space to $\C^{|G|}$, and recover real projectors by pairing conjugate irreps. Write $\rho_1,\dots,\rho_r$ for the irreducible representations, $d_i$ for their dimensions, and $\chi_i(g)=\operatorname{tr}\rho_i(g)$ for their characters, all $\rho_i$ taken unitary. Throughout we use
\begin{equation*}
L\;\cong\;\bigoplus_{i}d_i\,\rho_i,
\qquad
\sum_i d_i^2 \;=\; n .
\end{equation*}

\paragraph{The projector, entry by entry.} Since $[L_g]_{x,y}=1$ exactly when $g=xy^{-1}$, the operator of Eq.~\eqref{eq:idempotent} has entries
\begin{equation*}
[P_S]_{x,y}
\;=\;
\frac{1}{n}\sum_{i\in S} d_i\,\overline{\chi_i\!\left(xy^{-1}\right)} .
\end{equation*}
Once the conjugacy classes and the character table are available, each entry is read off the class of $xy^{-1}$. Nothing else is required, and in particular no data and no gradients. Lemma~\ref{lem:idem} records the properties used later.

\paragraph{The kernel.} The kernel $K_S=\bigcap_{i\in S}\ker\rho_i$ is an intersection of kernels of homomorphisms, hence a normal subgroup of $G$. Its left and right cosets therefore coincide, and $Q=G/K_S$ is a group. For a one-dimensional irrep the kernel is read directly from the character, $K=\{g:\chi(g)=1\}$. In the running example, $S_5$ has seven irreps, labeled by the partitions $(5)$, $(1^5)$, $(4,1)$, $(2,1^3)$, $(3,2)$, $(2^2,1)$ and $(3,1^2)$, of dimensions $1$, $1$, $4$, $4$, $5$, $5$ and $6$, and their squared dimensions sum to the order $120$. Only the trivial irrep $(5)$ and the sign irrep $(1^5)$ are unfaithful, with kernels $S_5$ and $A_5$. The two faithful irreps of least dimension, $(4,1)$ and $(2,1^3)$, are the two tied irreps, and either spans a minimal target subspace of dimension $16$ inside the $120$-dimensional input space.

\paragraph{The abelian reduction.} For $\Z_n$ the characters are $\chi_k(g)=\omega^{kg}$ with $\omega=e^{2\pi i/n}$ and all $d_k=1$, so
\begin{equation*}
P_{\{k\}}
\;=\;
\frac{1}{n}\sum_{g\in\Z_n}\omega^{-kg}\,L_g,
\end{equation*}
and the real projector attached to the conjugate pair $\{k,-k\}$ has image
\begin{equation*}
\operatorname{span}
\left\{
\left(\cos\tfrac{2\pi k x}{n}\right)_{x\in\Z_n},\;
\left(\sin\tfrac{2\pi k x}{n}\right)_{x\in\Z_n}
\right\},
\end{equation*}
the cosine and sine of a single frequency. The Fourier answer on modular addition is thus the abelian instance of Eq.~\eqref{eq:idempotent}. Moreover
\begin{equation*}
K_{\{k\}}=\{g\in\Z_n : kg\equiv 0 \bmod n\},
\end{equation*}
which is trivial exactly when $\gcd(k,n)=1$, so every coprime frequency is faithful on its own.

\paragraph{Completeness, quantified.} The completeness requirement of Section~\ref{sec:writing} rests on the max-margin readout of Eq.~\eqref{eq:readout}, which is bilinear in every matrix coefficient of each argument's irrep. A partial subspace of an irrep cannot express Eq.~\eqref{eq:readout}, because dropping coordinates of the contracted index changes the score even when the inputs remain distinguishable. The following proposition makes the loss quantitative for the score of the correct class. It concerns this readout and its margin, and the data-driven directions of Section~\ref{sec:supply} test it.

\begin{proposition}[Truncation cost]
\label{prop:margin}
Fix an irrep $\rho_i$, unitary of dimension $d_i$, and let $\Pi_J$ be the coordinate projector onto a
subset $J\subseteq\{1,\dots,d_i\}$ of the index contracted between the two arguments in
Eq.~\eqref{eq:readout}. If the embedding retains only the coordinates in $J$, the truncated
character term of the irrep at class $c$ is
$\operatorname{tr}\!\big(\rho_i(c)^{-1}\rho_i(a)\Pi_J\rho_i(b)\big)$. At the correct class
$c=ab$ its value is $|J|$, against $d_i$ when nothing is dropped.
\end{proposition}

\begin{proof}
Writing $\rho=\rho_i$ and using $\rho(ab)=\rho(a)\rho(b)$ with $\rho$ unitary,
\begin{align*}
\operatorname{tr}\!\big(\rho(ab)^{-1}\rho(a)\Pi_J\rho(b)\big)
&=\operatorname{tr}\!\big(\rho(b)^{-1}\rho(a)^{-1}\rho(a)\Pi_J\rho(b)\big)\\[2pt]
&=\operatorname{tr}\!\big(\rho(b)^{-1}\Pi_J\rho(b)\big)\\[2pt]
&=\operatorname{tr}\!\big(\Pi_J\,\rho(b)\rho(b)^{-1}\big)\\[2pt]
&=\operatorname{tr}\Pi_J \;=\; |J| .
\end{align*}
The second line uses $\rho(a)^{-1}\rho(a)=I$, the third the cyclicity of the
trace, and the fourth $\rho(b)\rho(b)^{-1}=I$. Taking $J=\{1,\dots,d_i\}$ gives $d_i$. Finally,
in the expansion
\begin{equation*}
\operatorname{tr}\!\big(\rho(c)^{-1}\rho(a)\rho(b)\big)
=\sum_{r,s,m}\overline{[\rho(c)]_{rs}}\,[\rho(a)]_{rm}\,[\rho(b)]_{ms},
\end{equation*}
every term with $m\notin J$ is absent from the truncated form, for all inputs.
\end{proof}

So a partial subspace does not merely lose precision. The correct-class contribution is scaled by $|J|/d_i$ while the deleted terms vanish identically. The proposition concerns this score, and Appendix~\ref{app:more} measures the corresponding test accuracy as coordinates are restored.

\section{Proofs}
\label{app:proofs}

The proofs are ordered by dependency. Lemmas~\ref{lem:idem} and~\ref{lem:onesided} supply the algebra behind Theorem~\ref{thm:unique}. Lemma~\ref{lem:collapse} turns the projector into a partition of the inputs, from which Theorem~\ref{thm:ceiling} and Corollary~\ref{cor:regular} follow. The remaining items explain why the closed form does not cover the squared conjugation, derive the tie-break law and the exact plateaus of Section~\ref{sec:corrupt} in that order, and give the first-passage form. There are no floats, so the derivations can be read straight through.

\subsection{The Idempotent Family}

The translations $L_g$ and $R_h$ are those of Eq.~\eqref{eq:commute}.

\begin{lemma}[Properties of the central idempotents]
\label{lem:idem}
The operators $P_i=\frac{d_i}{n}\sum_{g}\overline{\chi_i(g)}\,L_g$ satisfy
\begin{align*}
&\text{(i)}\quad P_i^{*}=P_i, \qquad P_i^2=P_i;\\
&\text{(ii)}\quad P_iP_j=0\ (i\neq j), \qquad \textstyle\sum_i P_i=I;\\
&\text{(iii)}\quad \operatorname{rank}P_i=d_i^2,
\qquad \operatorname{rank}P_S=\sum_{i\in S}d_i^2;\\
&\text{(iv)}\quad P_iL_g=L_gP_i,
\quad P_iR_h=R_hP_i
\quad(g,h\in G).
\end{align*}
\end{lemma}

\begin{proof}
\emph{(i) and (ii).} These are the Wedderburn decomposition of the group algebra
\citep{serre1977linear,fultonharris1991}. Under $\C[G]\cong\bigoplus_iM_{d_i}(\C)$ the element
$e_i=\frac{d_i}{n}\sum_g\overline{\chi_i(g)}\,g$ maps to the identity of the $i$-th summand and to
zero in the others, which gives idempotence, orthogonality and the fact that the $e_i$ sum to the identity. Self-adjointness follows
from $L_g^{*}=L_{g^{-1}}$ together with $\overline{\chi_i(g)}=\chi_i(g^{-1})$.

\emph{(iii).} The regular representation has
\begin{equation*}
\operatorname{tr}L_g=\#\{x\in G:\ gx=x\}=n\,\delta_{g,e},
\end{equation*}
so
\begin{align*}
\operatorname{tr}P_i
&=\frac{d_i}{n}\sum_{g}\overline{\chi_i(g)}\,\operatorname{tr}L_g\\
&=\frac{d_i}{n}\cdot\overline{\chi_i(e)}\cdot n\\
&=d_i\,\chi_i(e)\;=\;d_i^2 ,
\end{align*}
and a self-adjoint idempotent has rank equal to its trace. Additivity over $S$ follows from (ii).

\emph{(iv).} Left and right translations commute with each other, since
\begin{equation*}
R_h L_g\,e_x
=R_h e_{gx}
=e_{gxh}
=L_g\,e_{xh}
=L_g\,R_h\,e_x ,
\end{equation*}
so any linear combination of the $L_g$ commutes with every $R_h$. For the left translations,
conjugation-invariance of the character gives
\begin{align*}
L_h\,P_i\,L_h^{-1}
&=\frac{d_i}{n}\sum_{g}\overline{\chi_i(g)}\,L_{hgh^{-1}}\\
&=\frac{d_i}{n}\sum_{g'}\overline{\chi_i\!\left(h^{-1}g'h\right)}\,L_{g'}\\
&=\frac{d_i}{n}\sum_{g'}\overline{\chi_i(g')}\,L_{g'}
\;=\;P_i .
\end{align*}
\end{proof}

\begin{lemma}[One-sided invariance is not enough]
\label{lem:onesided}
Let $A=\{T:\ TR_h=R_hT\ \text{for all }h\in G\}$. Then
\begin{equation*}
A=\Big\{\textstyle\sum_{g}t(g)L_g\ :\ t\in\C^{G}\Big\}
\;\cong\;\bigoplus_i M_{d_i}(\C),
\end{equation*}
and an orthogonal projector in $A$ restricts on the $\rho_i$-isotypic component to an orthogonal
idempotent of $M_{d_i}(\C)$. Consequently $A$ contains projectors of rank $\sum_ik_id_i$ for every
choice $0\le k_i\le d_i$, and in particular, when $d_i>1$, projectors whose image is a proper subspace
of the $\rho_i$-isotypic component.
\end{lemma}

\begin{proof}
Let $TR_h=R_hT$ for all $h$ and put $t:=Te_e$. Since $e_x=R_xe_e$,
\begin{align*}
Te_x
&=TR_x e_e
=R_x Te_e
=R_x t\\
&=\sum_{y}t(y)\,e_{yx}
=\Big(\sum_{g}t(g)L_g\Big)e_x ,
\end{align*}
and as $x$ was arbitrary, $T=\sum_gt(g)L_g$. The reverse inclusion is the first computation in
Lemma~\ref{lem:idem}(iv). The map $t\mapsto\sum_gt(g)L_g$ is an algebra isomorphism from $\C[G]$. It is
multiplicative because $L_gL_h=L_{gh}$, and injective because the $L_g$ are linearly independent,
$L_ge_e=e_g$ being distinct basis vectors. Composing it with
$\C[G]\cong\bigoplus_iM_{d_i}(\C)$ gives the stated form. On the $\rho_i$-isotypic component, which
is isomorphic to $\C^{d_i}\otimes\C^{d_i}$ with $A$ acting on the first factor alone, an idempotent
of rank $k_i$ in $M_{d_i}(\C)$ has image of dimension $k_id_i$ inside a component of dimension
$d_i^2$, so any $0<k_i<d_i$ gives a proper subspace.
\end{proof}

Lemma~\ref{lem:onesided} is the precise content of the remark in Section~\ref{sec:construction} that one side alone would not suffice. Whenever $d_i>1$, right-invariance admits partial subspaces of the $\rho_i$-isotypic component, of every dimension that is a multiple of $d_i$, and it is the left-hand condition that removes them.

\subsection{Proof of Theorem~\ref{thm:unique}}

\begin{proof}
Let the orthogonal projector $P$ commute with every $R_h$ and every $L_g$. By
Lemma~\ref{lem:onesided}, $P=\sum_gt(g)L_g$ with $t=Pe_e$. Commuting with $L_h$,
\begin{align*}
P
&=L_h\,P\,L_h^{-1}\\
&=\sum_{g}t(g)\,L_{hgh^{-1}}\\
&=\sum_{g'}t\!\left(h^{-1}g'h\right)L_{g'},
\end{align*}
and linear independence of the $L_g$ forces
\begin{equation*}
t\!\left(h^{-1}gh\right)=t(g)
\qquad\text{for all }g,h\in G,
\end{equation*}
so $t$ is a class function. The irreducible characters are a basis of the class functions, so
\begin{equation*}
t=\sum_i c_i\,\frac{d_i}{n}\,\overline{\chi_i},
\qquad\text{that is}\qquad
P=\sum_i c_i P_i
\end{equation*}
for unique scalars $c_i$. By Lemma~\ref{lem:idem}(ii) the $P_i$ are orthogonal idempotents, hence
\begin{equation*}
P^2=\sum_i c_i^2\,P_i,
\end{equation*}
and $P^2=P$ forces $c_i^2=c_i$, so every $c_i\in\{0,1\}$. With $S=\{i:c_i=1\}$ we obtain $P=P_S$.
Conversely every $P_S$ is such a projector, by Lemma~\ref{lem:idem}. If $P$ is required to be real
then $t$ is real. Since $\overline{P_i}=P_{\bar\imath}$ for the conjugate irrep $\bar\imath$, $P_S$ is real
exactly when $S$ contains the conjugate of every irrep in $S$, which is the statement that complex
irreps enter in conjugate pairs.
\end{proof}

\subsection{Input Collapse}

\begin{lemma}[Collapse to cosets]
\label{lem:collapse}
Let the embedding of $x\in G$ be $P_Se_x$. Then
\begin{equation*}
P_Se_a=P_Se_{a'}
\quad\Longleftrightarrow\quad
a'a^{-1}\in K_S ,
\end{equation*}
so the embedding factors through $Q=G/K_S$ and is injective on $Q$.
\end{lemma}

\begin{proof}
Under $\C[G]\cong\bigoplus_iM_{d_i}(\C)$ the basis vector $e_x$ corresponds to the tuple
$(\rho_i(x))_i$, and $P_S$ deletes the coordinates outside $S$. Hence
\begin{align*}
P_Se_a=P_Se_{a'}
&\iff \rho_i(a)=\rho_i(a')\ \ \forall i\in S\\
&\iff \rho_i\!\left(a'a^{-1}\right)=I\ \ \forall i\in S\\
&\iff a'a^{-1}\in\bigcap_{i\in S}\ker\rho_i=K_S .
\end{align*}
Because $K_S$ is normal its right and left cosets coincide, and the induced map on $Q$ is injective
by construction.
\end{proof}

\subsection{Proof of Theorem~\ref{thm:ceiling}}

\begin{proof}
By Lemma~\ref{lem:collapse} the inputs reach the network only as the pair of cosets
$(\bar a,\bar b)\in Q^2$. Two input pairs in the same coset pair produce identical embedding vectors,
hence identical activations at every subsequent layer and identical logits, whatever the depth, the
width and the training time. A trained network therefore realizes a map $\phi:Q^2\to G$. For a set
$T$ of input pairs write
\begin{equation*}
m^{T}_{\bar a\bar b}(c)
=\#\big\{(a,b)\in T\cap(\bar a\times\bar b)\ :\ f(a,b)=c\big\},
\end{equation*}
so that the accuracy of $\phi$ on $T$ is
\begin{equation*}
\frac{1}{|T|}\sum_{(\bar a,\bar b)\in Q^2}
m^{T}_{\bar a\bar b}\big(\phi(\bar a,\bar b)\big).
\end{equation*}
This is maximized by choosing, independently in each coset pair, a class attaining
$\max_c m^{T}_{\bar a\bar b}(c)$. Taking $T$ to be all $|G|^2$ pairs and writing $m_{\bar a\bar b}$ for $m^{T}_{\bar a\bar b}$ gives the ceiling of Theorem~\ref{thm:ceiling},
\begin{equation}
\Acc_{\max}(S,f)=\frac{1}{|G|^{2}}\sum_{(\bar a,\bar b)\in Q^{2}}\ \max_{c\in G}\ m_{\bar a\bar b}(c).
\label{eq:ceiling}
\end{equation}
For a held-out split $T$ the same argument uses the counts $m^{T}$ and $|T|$ in the denominator. The
argument uses no property of the architecture beyond the factorization through $Q^2$, so it also
holds for the transformers of Section~\ref{sec:corrupt}, whose restricted token embeddings factor
through $Q$ in the same way.
\end{proof}

\subsection{Proof of Corollary~\ref{cor:regular}}

\begin{proof}
Let $w$ be an expression in which one argument occurs exactly once, and take it to be the second, so that
\begin{equation*}
w(a,b)=u(a)\,b^{\varepsilon}\,v(a),\qquad \varepsilon\in\{-1,1\},
\end{equation*}
for expressions $u,v$ in $a$ alone. Fix input cosets $A,B\in Q$. The quotient map $G\to Q$ is a
homomorphism, so every answer in the coset pair lies in the single coset $u(A)B^{\varepsilon}v(A)$,
of size $|K_S|$. For fixed $a\in A$ the map $b\mapsto u(a)b^{\varepsilon}v(a)$ is a bijection from
$B$ onto that coset, since left and right multiplication are bijections, as is inversion when
$\varepsilon=-1$. Each attainable answer therefore occurs exactly once for every $a\in A$, that is
$|K_S|$ times among the $|K_S|^2$ pairs of the coset pair, so $\max_cm_{\bar a\bar b}(c)=|K_S|$ for
every coset pair. The same argument applies with the two arguments exchanged. Hence
\begin{align*}
\Acc_{\max}
&=\frac{1}{|G|^2}\,|Q|^2\,|K_S|\\[2pt]
&=\frac{1}{|G|^2}\cdot\frac{|G|^2}{|K_S|^2}\cdot|K_S|
\;=\;\frac{1}{|K_S|}.
\end{align*}
Multiplication $ab$ and conjugation $aba^{-1}$ both satisfy the hypothesis.
\end{proof}

\begin{remark}[Expressions with no once-occurring argument]
\label{rem:squared}
For the squared conjugation $ab^2a^{-1}$ the relevant map on $K_S$ is
$l\mapsto l\cdot\big(b_0lb_0^{-1}\big)$, which need not be injective. The attainable answers within a
coset pair can then be non-uniformly distributed and $\max_cm_{\bar a\bar b}(c)$ can exceed $|K_S|$. When it
does, the ceiling exceeds $1/|K_S|$. It remains computable from Eq.~\eqref{eq:ceiling}, which is
how the sweep of Appendix~\ref{app:details} evaluates it, and only the closed form $1/|K_S|$ is lost.
\end{remark}

\subsection{The Tie-Break Law}

\begin{proposition}
\label{prop:tiebreak}
If the output weights are projected as well, the logit vector lies in the image of $P_S$ and is
therefore constant on the cosets of $K_S$, so any two classes in the same coset receive equal logits
at every step of training. Suppose that the answer coset of each pair has a strictly higher score
than every other coset. The prediction on each pair is then the class that the implementation's
tie-break selects in its answer coset, so the realized plateau is a deterministic function of the
split and of that rule, independent of the initialization, and is bounded above by
Eq.~\eqref{eq:ceiling} restricted to the split.
\end{proposition}

\begin{proof}
By Lemma~\ref{lem:collapse}, $P_Se_c=P_Se_{c'}$ for classes $c,c'$ in the same coset, so for a logit
vector $z=P_Sz$ self-adjointness gives $e_c^\top z=(P_Se_c)^\top z=(P_Se_{c'})^\top z=e_{c'}^\top z$,
and the projection enforces this equality at every forward pass. Under the hypothesis the
highest-scoring coset of a pair is its answer coset, and the tie-break picks the same representative
in it for every seed, so a test pair is correct exactly when its answer is that representative.
Counting these pairs on $T$ gives the realized plateau, and Theorem~\ref{thm:ceiling} applied to $T$,
which maximizes over the choice of representative, bounds it.
\end{proof}

\subsection{The Two Nested Restrictions of Section~\ref{sec:corrupt}}
\label{app:exact}

Section~\ref{sec:corrupt} reports two configurations of $S_5$ multiplication whose plateaus are hit exactly by every seed. Sixteen further runs on the same split, three per configuration from the split sweep and five from a separate repeat, land on the same two counts. The split holds a fraction $0.4$ of the pairs for training, so
\begin{equation*}
|G|^2=14{,}400,
\qquad
|T|=0.6\cdot 14{,}400=8{,}640 .
\end{equation*}
Ablating every faithful irrep leaves the trivial and sign irreps, whose kernels are $S_5$ and $A_5$, so
\begin{align*}
K_S&=S_5\cap A_5=A_5, & |K_S|&=60, & |Q|&=2,
\end{align*}
giving four coset pairs of $3{,}600$ pairs each in which, by Corollary~\ref{cor:regular}, the answers are the $60$ elements of one coset, each occurring $60$ times. Ablating the sign irrep as well leaves the trivial irrep alone, so
\begin{align*}
K_S&=S_5, & |K_S|&=120, & |Q|&=1,
\end{align*}
a single coset pair in which each of the $120$ classes occurs $120$ times. Over all pairs Corollary~\ref{cor:regular} therefore gives $1/60$ and $1/120$, levels that do not depend on the split.

Three levels have to be kept apart on a given split. The uniform level of Corollary~\ref{cor:regular} is
\begin{align*}
\tfrac{1}{60}=\tfrac{144}{8640}=0.016667,
\qquad
\tfrac{1}{120}=\tfrac{72}{8640}=0.008333 .
\end{align*}
The split-restricted bound of Theorem~\ref{thm:ceiling}, obtained by taking the modal answer in each coset pair of the test split, is larger. For the four coset pairs of the first configuration the maxima are $43$, $46$, $45$ and $46$, so
\begin{align*}
\text{first:}&\quad \tfrac{43+46+45+46}{8640}=\tfrac{180}{8640}=0.020833,\\[2pt]
\text{second:}&\quad \tfrac{87}{8640}=0.010069 .
\end{align*}
The realized plateaus, which lie between the two levels and coincide with neither, are
\begin{align*}
\tfrac{149}{8640}=0.017245,
\qquad
\tfrac{75}{8640}=0.008681 ,
\end{align*}
each attained by all ten seeds with zero variance. They are the counts selected by the tie-break rule of Proposition~\ref{prop:tiebreak}, and not the modal counts of Theorem~\ref{thm:ceiling}, which the network does not reach because projecting the output weights ties the scores of the classes inside each coset.

All three levels can be computed on any split, and the realized plateau follows the tie-break count each time. At the $30\%$ training split of the sweep, where $|T|=10{,}080$, the tie-break counts are $170$ and $86$, the modal bounds are $198$ and $97$, and the observed plateaus are $170/10080$ and $86/10080$. On four further random splits with $40\%$ of the pairs for training, the counts move with the split, to $146$ and $71$, $155$ and $70$, $167$ and $75$, and $144$ and $70$, and the trained networks land on them exactly, with zero variance across the three seeds of each configuration.

\subsection{The First-Passage Form}

The form expected in Section~\ref{sec:predictions} comes from the simplest dynamics in which a head start compounds.

\begin{proposition}[\textbf{First-passage form}]
\label{prop:passage}
Suppose that during the delay the alignment $\alpha(t)$ of the embedding with the target subspace grows self-reinforcingly,
\begin{equation*}
\dot\alpha=\gamma\,\alpha^{\,1+p}
\qquad\text{with }\gamma,p>0,
\end{equation*}
and that grokking occurs a fixed consolidation time $t_w$ after $\alpha$ first reaches a threshold $\alpha_\star\le1$. Then every run with initial alignment $0<\alpha_0<\alpha_\star$ groks at the $\tau(\alpha_0)$ of Eq.~\eqref{eq:alignlaw}, with $C=1/(\gamma p)$ and $t_0=t_w-C\,\alpha_\star^{-p}$, and every run with $\alpha_0\ge\alpha_\star$ groks within the consolidation time.
\end{proposition}

\begin{proof}
Separating variables in the growth law above,
\begin{align*}
\alpha^{-1-p}\,d\alpha &= \gamma\,dt,\\[2pt]
-\tfrac{1}{p}\,\alpha^{-p} &= \gamma t + \text{const},\\[2pt]
\alpha(t)^{-p} &= \alpha_0^{-p}-p\gamma t ,
\end{align*}
using $\alpha(0)=\alpha_0$. The threshold $\alpha_\star$ is first reached at
\begin{equation*}
t_\star=\frac{\alpha_0^{-p}-\alpha_\star^{-p}}{p\gamma},
\end{equation*}
which is finite and positive exactly when $0<\alpha_0<\alpha_\star$. Adding the consolidation time
$t_w$,
\begin{align*}
\tau(\alpha_0)
&=t_w+\frac{\alpha_0^{-p}-\alpha_\star^{-p}}{p\gamma}\\[2pt]
&=\underbrace{\left(t_w-\frac{\alpha_\star^{-p}}{p\gamma}\right)}_{t_0}
\;+\;\underbrace{\frac{1}{p\gamma}}_{C}\;\alpha_0^{-p},
\end{align*}
which is Eq.~\eqref{eq:alignlaw} with $C=1/(\gamma p)$ and $t_0=t_w-C\alpha_\star^{-p}$. For
$\alpha_0\ge\alpha_\star$ the threshold is met at $t=0$ and grokking occurs within $t_w$.
\end{proof}

\begin{remark}[Scope of the model]
The solution diverges in finite time, so the growth law above is meaningful only up to the threshold. It is also monotone by construction, whereas measured alignment trajectories need not be monotone (Figure~\ref{fig:conc}). Eq.~\eqref{eq:alignlaw} is therefore a hitting-time relation between $\tau$ and $\alpha_0$, not a description of the path.
\end{remark}

First-passage arguments have also been applied to the weight norm~\citep{khanh2026firstpassage}, where the delay is modeled as the time for the squared parameter norm to reach a threshold calibrated for each architecture. Here the quantity that reaches a threshold is the alignment with a subspace computed before training. The two views are compatible in our runs, where weight decay changes the speed of the search but not the dependence of $\tau$ on $\alpha_0$ (Appendix~\ref{app:dynamics}).

\subsection{Restriction Behind an Encoder}

\begin{proposition}
\label{prop:gequiv}
Suppose the group acts on the inputs only through an encoder $\psi$ that is not equivariant, so that
the features $\psi(x)$ do not carry the regular representation, and write
$v=(P_S\psi(x_1),P_S\psi(x_2))$ for the projected features of an input pair. Then $P_S$ projects onto
a subspace of feature space that need not contain the solution of Eq.~\eqref{eq:readout}, the
hypothesis of Lemma~\ref{lem:collapse} fails, and on any evaluation set $T$ the attainable accuracy
is bounded not by Eq.~\eqref{eq:ceiling} but by the collision bound of the composite map,
\begin{equation}
\label{eq:encbound}
\Acc_{\max}
\;\le\;
\frac{1}{|T|}\sum_{v\,\in\,V_T}\;\max_{c}\;m^{T}_{v}(c),
\end{equation}
where $V_T$ is the set of projected feature pairs on $T$ and $m^{T}_{v}(c)$ counts the pairs in $T$
with projected features $v$ and answer $c$. The sum runs over the fibres of the composite map rather
than over pairs of cosets of $K_S$, so it depends on the encoder, and it is in general neither
$1/|K_S|$ nor attained.
\end{proposition}

\begin{proof}
Lemma~\ref{lem:collapse} uses only one property of the restriction. Two inputs that receive
the same vector after the projection receive the same vector at every later layer, so the network's
answer on them agrees and the best it can do is bet, per fibre, on the modal answer. That argument
is purely about the partition of the input pairs induced by the map that feeds the network, and it
applies verbatim to the composite map, which gives Eq.~\eqref{eq:encbound}.

What equivariance supplies is the \emph{identity} of that partition. For one-hot group inputs,
Lemma~\ref{lem:collapse} identifies the fibres with pairs of cosets of $K_S$, all of the same size,
and over all pairs Eq.~\eqref{eq:encbound} reduces to the uniform level $1/|K_S|$ of
Corollary~\ref{cor:regular} for tasks with a once-occurring argument. Without equivariance $\psi$
need not commute with the action, so nothing forces the fibres to be unions of coset pairs, to be
equally sized, or to align with the labels. The image of $P_S$ also need not contain a solution of
Eq.~\eqref{eq:readout}, so the bound need not be attained either. The projection can only merge
inputs that the unrestricted encoder separated, so it cannot raise the bound, while trained
classifiers may fall below their bounds by different amounts.

The prediction is therefore a loss rather than a level, and it is a loss that carries no signature of
the subspace, because nothing in the argument distinguishes $P_S$ from any other projector of its rank once
the fibres stop being cosets.
\end{proof}

\section{Additional Experiments and Controls}
\label{app:more}

This section settles three questions in the order in which the paper raises them, namely what else the subspace could have been, how restricting to it compares with methods that change training instead, and what the ceiling does not depend on.

\subsection{Alternatives to the Target Subspace}

\paragraph{Alternative subspaces.} A uniformly random $16$-dimensional subspace, which lies outside the family $\{P_S\}$ and so outside the reach of Theorem~\ref{thm:ceiling}, never fits the training set, and its test accuracy stays at $\le.0082$, below the chance level $1/120=.0083$. Each competing description of the destination was run at its own rank or at matched alignment. The reverse-engineered coset basis and half of the target subspace never grok, and neither fits the training set. The $S_4$ coset basis caps train and test accuracy together at $.017$, the two-dimensional basis of the $A_5$ cosets stops at $.016$ and $.017$, and half of the $(4,1)$ subspace caps train at $.36$ and test at $.24$, values that it reaches by step $3{,}850$ and holds to the end of its $10$k-step runs. The top sixteen principal directions of a converged network cross the $0.95$ threshold between $2{,}500$ and $9{,}500$ steps, but their best test accuracy over the $100$k-step runs stays between $.967$ and $.970$. Full accuracy returns only with the missing coordinates, at rank $26$ ($.9995$ to $1.000$, median grokking step $500$) and rank $41$ ($.9997$ to $1.000$, median grokking step $1{,}000$). Proposition~\ref{prop:margin} gives the cost in the character score. Dropping coordinates of an irrep scales the correct-class term by the fraction of coordinates retained. The coset construction is not itself at fault. Completed to the two irreps that the $S_4$ coset space carries, the trivial irrep and $(4,1)$, it becomes a rank-$17$ subspace that contains the whole minimal target subspace, and it groks at $150$ steps on ten of ten seeds, exactly like the target subspace. What fails is the five-dimensional basis of coset indicators alone. A random rank-$5$ subspace, run on the same ten seeds, fits neither set, with train accuracy at most $.041$ and test accuracy at most $.010$.

\paragraph{Rank is not the ingredient.} Freely trained rank-$k$ factorizations $E=AB$ never beat vanilla training. At ranks $16$, $25$ and $41$, with the output weights factorized at the same rank, none of five runs groks within $90$k steps, although the rank-$41$ runs fit the training set. At rank $48$, one of five such runs groks, and two of five with the output weights left free, at a median of $73{,}850$ steps over the three. Within the same $90$k steps, $16$ of the $20$ vanilla runs grok. Together with the principal directions, the free ladder separates two kinds of failure. Training low-rank embeddings without confining them to the family fails or is at best no faster than vanilla training, while a partial subspace, which violates completeness, caps the accuracy, and the cap falls as more of the target subspace is missing.

\subsection{Methods That Change Training Instead}

\paragraph{Training-side accelerators.} Six published accelerators, SAM~\citep{foret2021sharpness}, Muon~\citep{tveit2025muon}, Grokfast~\citep{lee2024grokfast}, Hessian-aware noise~\citep{zhang2024hessian}, GrokAlign~\citep{walker2025grokalign} and embedding transfer~\citep{xu2025groktransfer}, were grid-searched under one protocol and compared against the target subspace, which uses oracle information and serves as a diagnostic rather than a proposal. Table~\ref{tab:accel} reports the best configuration of each, selected by the number of runs that grok within $90$k steps and then by the median. Muon fails to grok multiplication at any grid point. Of the five that do, four land within a factor of three of vanilla training, and the best, GrokAlign, is $14\times$ faster than vanilla training on the same three seeds and still $28\times$ slower than the target subspace. Embedding transfer is counted from the step at which the precursor is loaded, without the cost of training the precursor. 

\begin{table}[htbp]
\centering\footnotesize
\caption{\textbf{Training-side accelerators on $S_5$ multiplication.} Each accelerator ran on seeds $0$, $1$ and $42$ at its best grid point. \emph{Runs} counts the runs that grok within $90$k steps ($200$k for the primary set), and $\tau$ is their median grokking step.}\label{tab:accel}
\setlength{\tabcolsep}{4pt}
\begin{tabular*}{\columnwidth}{@{\extracolsep{\fill}}llcr@{}}
\toprule
Method & Configuration & Runs & $\tau$ \\
\midrule
Vanilla (primary set) & --- & $20/20$ & $64{,}050$ \\
Vanilla (same seeds) & --- & $3/3$ & $59{,}250$ \\
\midrule
SAM & $\rho=0.05$ & $3/3$ & $60{,}500$ \\
Muon & lr $0.1$, $m=0.99$ & $0/3$ & --- \\
Grokfast & $\lambda=1$ & $3/3$ & $77{,}500$ \\
Hessian-aware noise & $\sigma=0.03$ & $3/3$ & $21{,}500$ \\
GrokAlign & default & $3/3$ & $4{,}250$ \\
Embedding transfer & rank-$64$ precursor & $3/3$ & $42{,}800$ \\
\midrule
Target subspace & --- & $10/10$ & $150$ \\
\bottomrule
\end{tabular*}
\end{table}

\subsection{What the Ceiling Does Not Depend On}

\paragraph{Robustness of the ceilings.} The ceiling depends on what the input projection lets any architecture see, and on nothing else. We test two of the things that it should not depend on, the amount of label noise and the depth of what follows the embedding. With up to $30\%$ of training labels resampled uniformly, the sign-irrep plateau stays at $.0172$, bit-identical across the five seeds of each setting, against a computed all-pairs ceiling of $.0167$, and the trivial irrep stays at $.0087$ against $.0083$. The restriction to the target subspace still groks at every noise level, at a median of $150$ steps up to $20\%$ and $200$ steps at $30\%$. A six-layer transformer reaches the sign-irrep plateau of the one-hidden-layer network, $149/8640$, on all five seeds within $3{,}000$ steps and never passes it, although three of the five fall below it at some later evaluations and one ends at $75/8640$. The restriction to the target subspace groks at every depth tested, at median steps of $150$, $2{,}000$, $2{,}000$, $500$, $1{,}500$ and $5{,}000$ for one, two, three, six, eight and twelve layers. The twelve-layer runs grok between $2{,}500$ and $9{,}000$ steps, and the one seed stopped at $6$k steps before grokking groks at $7{,}000$ when rerun with a budget of $15$k. These transformer runs come from an exploratory sweep over depth, learning rate and weight decay, with weight decay $1$ throughout and learning rate $10^{-3}$ at two and three layers and $3\cdot10^{-4}$ from six layers on. Vanilla transformers at the same settings grok far later or not at all. At two layers, two of five runs grok within $60$k steps, at $23{,}250$ and $25{,}750$. At six layers, six of seven grok, between $19{,}000$ and $41{,}000$ steps, and the seventh stops at $31{,}500$ steps without grokking. At eight layers none of four groks within $40$k steps, and at twelve layers two of six grok, at $16{,}000$ and $24{,}000$, while the other four stop between $2{,}500$ and $30{,}000$ steps. The sweep has one pair in which vanilla training grokked first, a single six-layer seed at learning rate $10^{-3}$, where the vanilla run groks at $5{,}500$ steps and the restricted one at $12{,}500$. Four further six-layer vanilla runs at that learning rate, with weight decays from $0.03$ to $0.5$, stop at $26{,}500$ to $28{,}500$ steps without grokking, and six short pilot runs restricted to the sign or the trivial irrep never pass their plateaus. The computed ceiling is unchanged by construction, because Theorem~\ref{thm:ceiling} constrains the input pathway only.

\section{Experimental Details}
\label{app:details}

The protocol first, then details of the experiments of Section~\ref{sec:experiments} in the order the paper uses them.

\subsection{Protocol}

\paragraph{Protocol.} Unless noted, the network is Eq.~\eqref{eq:net} with embedding dimension $256$ and width $128$, trained full-batch with AdamW at learning rate $10^{-3}$, decoupled weight decay $1.0$ and $\beta=(0.9,0.98)$, on a $40\%$ training split of $S_5$ multiplication with a fixed data seed. Restrictions apply $P_S$ to both embeddings and to the output weights at every forward pass, except in the commutator and squared-conjugation configurations of the ceiling sweep, which project the inputs only. The initialization-only variant projects all three matrices once, at initialization, and never again. Every restricted initialization is scaled by $\sqrt{|G|/\operatorname{rank}P_S}$, which keeps its expected squared norm equal to that of an unrestricted one. The grokking step is $\tau=\min\{t:\mathrm{Acc}_{\mathrm{test}}(t)\ge0.95\}$ and the memorization step is the first step at which training accuracy reaches $1.0$. The primary set is the twenty vanilla seeds of Figure~\ref{fig:delay}. Budgets differ by experiment. They are $200$k for the primary set, $10$k for the restricted setting, $500$k for the rotation grid and its random-subspace control, and $10$k to $100$k for the alternative subspaces. Other budgets are given where the experiment is described. The code, configurations and logs of every experiment are in the supplementary material.

\subsection{Restricting to the Target Subspace}

\paragraph{Across data splits.} Every headline number of Section~\ref{sec:experiments} comes from one train--test partition, so the primary settings were repeated on five splits. Vanilla runs use ten seeds per split, the restricted setting five, and the rotated settings three per level.

On splits $42$, $0$, $1$, $7$ and $13$, the vanilla ten-seed medians are $76{,}650$, $84{,}850$, $83{,}200$, $74{,}150$ and $73{,}200$ steps, and all ten seeds grok on every split. The restricted median is $150$ on every split, individual runs falling between $140$ and $160$. The split moves the starting point and not the destination. The dependence of $\tau$ on $\alpha_0$ is monotone on all five, with the $\alpha_0=0.4$ level between $1{,}760$ and $2{,}600$ steps, and the fitted exponents are $3.39$, $3.71$, $3.50$, $4.18$ and $3.59$. On split $42$, the split of the main text, this smaller design gives $3.39$, inside the $95\%$ interval $[2.98,3.49]$ of the $85$-run fit. The spread across splits is a further reason to report the form rather than the exponent. The vanilla figures here are ten-seed medians and are not the twenty-seed median of $64{,}050$ quoted in the main text. On the split of Section~\ref{sec:prelim} the first ten of those twenty seeds give $76{,}650$.

\paragraph{Weight decay and the projection.} Without weight decay, on seeds $0$ to $3$ and $42$ with a budget of $100$k steps, the restricted network groks at step $200$ on all five seeds, and a network projected once at initialization and never again groks at step $300$ on all five. Vanilla training without weight decay memorizes by step $400$, but its test accuracy never exceeds $.0081$, against a chance level of $.0083$, while $16$ of the $20$ vanilla runs with weight decay grok within the same $100$k steps. The soft penalty, at strengths $0.3$, $1$, $3$ and $10$ with five seeds each, groks between $500$ and $800$ steps in all twenty runs, and with weight decay the network projected once at initialization groks between $240$ and $270$ steps on all five seeds.

\paragraph{Alignment grids and model selection.} On $S_5$ with the winning irrep $(4,1)$ as target, thirteen alignment levels spanning $[0.14,0.995]$ were run at five seeds each, rotating the two input embeddings of the random initialization toward the target subspace while preserving their norms, with the output weights left as initialized, and every level groks. Pooling the sixty-five rotated runs with the twenty vanilla runs gives the fit reported in Section~\ref{sec:supply}, with the random-subspace control excluded from the fit. The five candidate forms, fixed before fitting, were compared by least squares on $\log\tau$ over the $85$ fitted runs, and the expected form wins decisively. The power law with a floor, $t_0+C\alpha_0^{-p}$, the only form with three free parameters, has an AIC of $-179.8$, against $-147.2$ for the pure power law $C\alpha_0^{-p}$, $-46.2$ for the exponential form $Ae^{-b\alpha_0}$, $+19.3$ for the logarithmic form $A-b\log\alpha_0$ and $+120.4$ for the hyperbolic form $A/(c-\alpha_0)$. The selected form gives $p=3.22$ $[2.98,3.49]$, $t_0=459$ $[393,522]$ and $C=99.5$ $[66.1,140.3]$, with $95\%$ intervals from $10{,}000$ bootstrap resamples over runs.

Two further checks bear on the selected form.

\begin{enumerate}[leftmargin=1.2em,itemsep=2pt]
\item Refitting on the sixty-five rotated runs alone gives $p=3.163$, $t_0=447$ and $C=107$. Evaluated at the vanilla runs' median initial alignment $\alpha_0=0.1333$,
this refit predicts a grokking step of $63{,}158$ against an observed median of $64{,}050$, a ratio of
$0.986$. The vanilla runs enter the check only as the quantity predicted.
\item Three checks on the selection itself. Refitting each rejected form on the sixty-five rotated
runs alone and evaluating it at the vanilla runs' median $\alpha_0$ gives $45{,}194$, $21{,}230$,
$11{,}721$ and $3{,}347$ against the observed $64{,}050$, errors of $29\%$, $67\%$, $82\%$ and $95\%$,
against the selected form's $1.4\%$. The decreasing hyperbola $A/(\alpha_0-c)$, whose pole lies on the other side of the data, predicts $726{,}952$, a factor of eleven too high. Adding a floor to the exponential and logarithmic forms and to the decreasing hyperbola,
so that the comparison is matched at three free parameters, leaves the margin essentially unchanged
at $\Delta\mathrm{AIC}=30.2$ to the nearest floored competitor, the exponential form, against $32.6$ to the pure power law without floors, so the form
does not win by having one. Resampling the fourteen design points rather than the eighty-five runs,
which is the level at which the design is replicated, widens the interval on $p$ only from
$[2.98,3.49]$ to $[2.88,3.46]$.
\end{enumerate}

\paragraph{The random-subspace control.} The control applies the same norm-preserving rotation toward a random rank-$16$ subspace, drawn afresh for every run, over the same thirteen alignment levels and five seeds as the treatment. All sixty-five runs grok, and none is censored. The grokking step does not depend on the alignment. The slope of $\log\tau$ on $\log\alpha_0$ is $+0.071$ with a bootstrap interval of $[-0.066,+0.175]$ over the design points, so the exponent of Eq.~\eqref{eq:alignlaw} is not identified at all for this arm, and several levels are slower than vanilla training. Two checks confirm the rotation did what it was asked to. The alignment with each run's own random subspace follows the nominal level to three decimals, and the median alignment with the target subspace stays between $0.129$ and $0.134$ at every level, the chance alignment. Concentrating the embedding in sixteen directions therefore buys nothing unless they span the target subspace.

\paragraph{Across optimizers.} The role of the target subspace was re-checked under two further optimizers, with two choices that matter for reading Table~\ref{tab:optimizers}. First, weight decay is held at the protocol's value in every arm, so that the optimizer is the only difference. Second, the mini-batch arm changes the number of updates per epoch as well as the batch size, so the comparison is reported in steps and in epochs. Under both optimizers vanilla training memorizes and generalizes almost together, leaving no delay to shorten. The target subspace nevertheless brings generalization far sooner. More alignment still helps monotonically, with median grokking steps of $770$ and $1{,}460$ at $\alpha_0=0.9$ and $0.5$ under mini-batch AdamW and $3{,}200$ and $11{,}000$ under SGD, three seeds each, all grokking. Under mini-batch AdamW the unfaithful setting still lands on the value computed in Appendix~\ref{app:exact}, $149/8640$ with zero variance, and none of its three runs groks. In epochs of twelve optimizer steps, mini-batch AdamW memorizes at epoch $1{,}267$ and groks at epoch $1{,}608$, and its restricted network groks at epoch $22$. The search is therefore $39.8$ times faster than under full-batch training while memorization is $3.25$ times slower, which is the sense in which the delay appears where the search is slow relative to memorization.

\begin{table}[htbp]
\centering\footnotesize
\caption{\textbf{The delay belongs to the protocol, but the target subspace does not.} Median memorization step and grokking step $\tau$ in optimizer steps, over twenty vanilla and ten restricted runs for full-batch AdamW and five each for the other optimizers. Every run grokked.}\label{tab:optimizers}
\setlength{\tabcolsep}{4pt}
\begin{tabular*}{\columnwidth}{@{\extracolsep{\fill}}lrrrr@{}}
\toprule
 & \multicolumn{2}{c}{Vanilla} & & \\
\cmidrule(lr){2-3}
Optimizer & Mem.\ step & $\tau$ & Restr.\ $\tau$ & Speedup \\
\midrule
Full-batch AdamW & $390$ & $64{,}050$ & $150$ & $427\times$ \\
Mini-batch AdamW & $15{,}200$ & $19{,}300$ & $260$ & $74\times$ \\
SGD with momentum & $21{,}600$ & $23{,}200$ & $1{,}070$ & $22\times$ \\
\bottomrule
\end{tabular*}
\end{table}

\subsection{Ablating the Target Subspace}

\paragraph{Ceiling sweep.} The $163$ configurations comprise a dense multiplication sweep ($108$), conjugation ($29$), the commutator ($13$) and the squared conjugation ($13$), over seventeen groups, which are six cyclic groups ($\Z_{60}$, $\Z_{72}$, $\Z_{90}$, $\Z_{120}$, $\Z_{144}$, $\Z_{180}$), six dihedral groups ($D_8$, $D_{10}$, $D_{12}$, $D_{15}$, $D_{20}$, $D_{30}$), and $A_5$, $S_5$, $A_6$, $S_6$ and $\mathrm{PSL}(2,7)$. Seventy-nine of the $108$ multiplication configurations are abelian, one per distinct $\gcd(n,k)$ of a single character. For each multiplication and conjugation configuration the ceiling is computed before training as the uniform level $1/|K_S|$ of Corollary~\ref{cor:regular}, which does not depend on the split. For the commutator and the squared conjugation, which keep only the trivial input irrep, it is the frequency of the most common answer over all pairs. The plateau is the median of the last six test evaluations, taken every $250$ steps, averaged over the seeds of a configuration where there are three. Six configurations have a trivial kernel, and the mean relative deviation is taken over the other $157$. The $108$ multiplication and $29$ conjugation configurations project the output weights as well as the embeddings, while the $26$ commutator and squared-conjugation configurations project the input only, since their level is a global majority for which the output weights play no role. The $137$ multiplication and conjugation configurations are single-seed by design, because both routes to the plateau are deterministic, the projected one by Proposition~\ref{prop:tiebreak} and the collapsed one because a single coset leaves one available answer. The $26$ commutator and squared-conjugation configurations use three seeds each and confirm zero variance. The sweep network has embedding dimension $128$, or the protocol's $256$ where only the input is projected. The multiplication sweep trains on $30\%$ of the pairs and the other tasks on $40\%$, with at most $12{,}000$ training pairs, while the two configurations of Appendix~\ref{app:exact} run at the protocol's $40\%$. For those two restrictions, the ratio of the plateau to the all-pairs ceiling is $1.012$ and $1.024$ in the sweep, against $1.035$ and $1.042$ under the protocol. 

\subsection{Observing Vanilla Training}

\paragraph{Circuit probe.} At a probe step the logits $z_c(a,b)$ are centered per input pair and fitted by least squares as in Eq.~\eqref{eq:decomposition}. Write $z^{(i)}$ for the term of irrep $i$, $z^{\mathrm{gen}}=\sum_iz^{(i)}$ for the generalizing component, and $r=z-z^{\mathrm{gen}}$ for the memorizing component, all taken over every input pair and class. The generalizing curve in the right panel of Figure~\ref{fig:anatomy} is $\|z^{\mathrm{gen}}\|^2/\|z\|^2$, the memorizing curve is the argmax accuracy of $r$ over all input pairs, and the share of irrep $i$ is $\|z^{(i)}\|^2/\|z^{\mathrm{gen}}\|^2$. Probes are taken every $50$ steps to $2{,}000$ and every $500$ steps thereafter to $200$k, $437$ probe steps per run. Every curve in Figure~\ref{fig:search} is the median across the twenty runs at each probe step. The sign irrep holds a share of $0.96$ at step $10$k, and at step $45$k the memorizing component reaches a median accuracy of $0.019$ while the median test accuracy is $0.025$.

\begin{figure}[htbp]\centering
\includegraphics[width=\columnwidth]{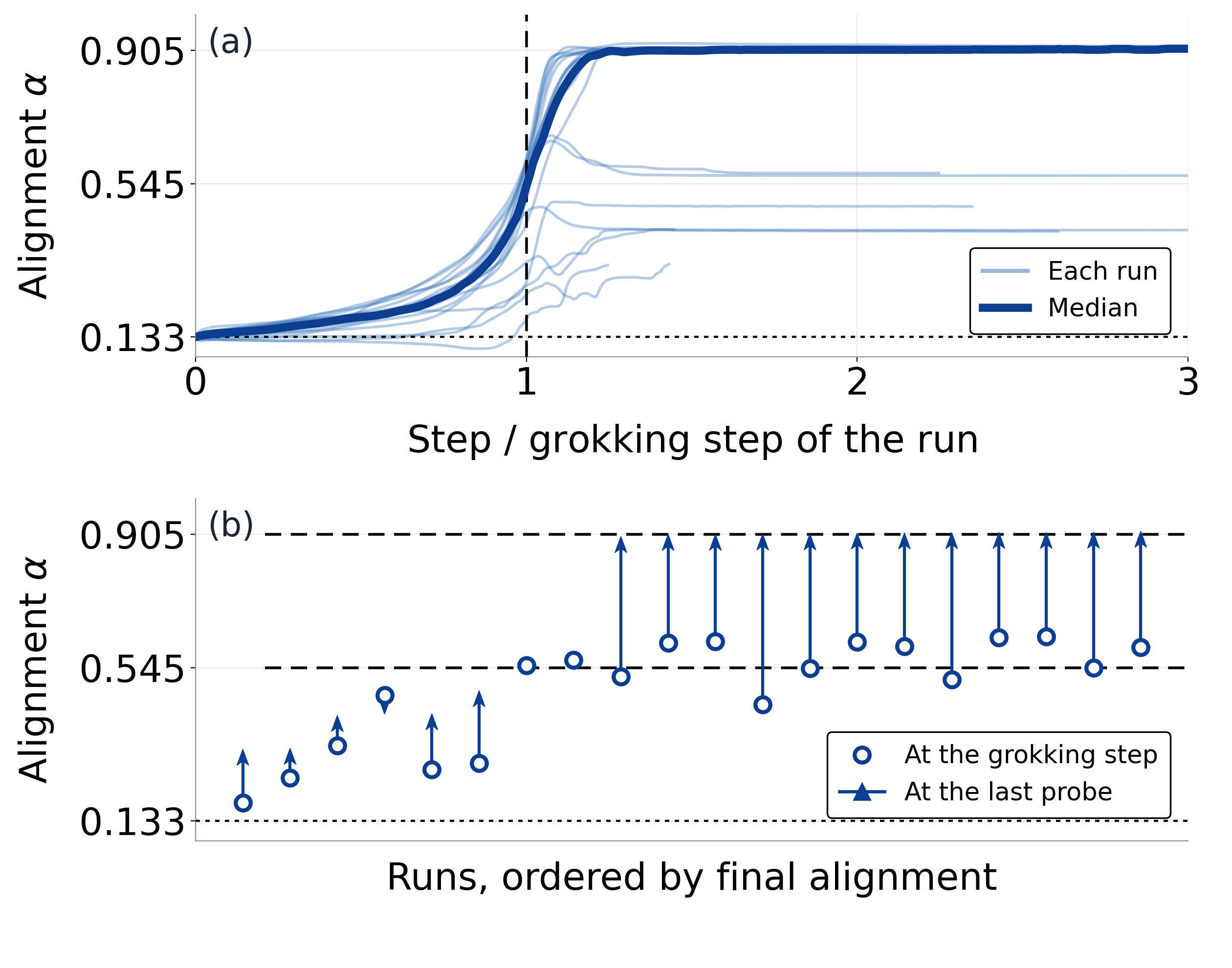}
\caption{\textbf{Nobody told these runs where to go.} \textbf{(a)} Alignment with the target subspace
against time normalized by each run's grokking step. \textbf{(b)} Alignment at the first probe at or after the grokking step
(dot) and at the last probe (arrowhead), one segment per run. Generalization begins once a run is close
enough, not after it has settled.}
\label{fig:conc}
\end{figure}

\paragraph{Alignment without a prior.} With fully free embeddings and no projection anywhere, the alignment of the converged embedding with the target subspace has median $0.905$ over the twenty runs, against $0.133$ for a random subspace of the same dimension. At the first probe at or after each run's grokking step the median is $0.545$ and still rising. Twelve of twenty runs exceed $0.85$ at convergence, while eight settle between $0.327$ and $0.574$ and grok regardless, and the two least aligned runs are the two slowest, at $0.327$ with $\tau=160{,}500$ and $0.329$ with $\tau=139{,}800$. The projector is used here only to measure. Figure~\ref{fig:conc} plots every run.

\paragraph{Steering the tie.} The steering arm plants the initialization advantage on the other tied irrep, with ten seeds at $\alpha_0=0.15$ and five each at $0.165$ and $0.18$. The flip fraction rises with the planted alignment, from $3/10$ at $0.15$ to $4/5$ at $0.165$ and $5/5$ at $0.18$. One of the seven unflipped runs at $0.15$ was censored at the $200$k cap, and the fraction $3/10$ keeps it in the denominator.

\subsection{Intervening During Training}
\label{app:dynamics}

\paragraph{Rotation in mid-run.} Five vanilla runs of the primary protocol, seeds $0$ to $4$, were saved at $t^{\ast}=5$k and at $t^{\ast}=30$k and resumed with both embeddings rotated to an alignment of $0.4$ or $0.7$ with the minimal target subspace, or down to $0.05$, below the chance alignment, all norms preserved and the optimizer state carried over, or resumed without rotation, under a budget of $500$k steps. Resuming without rotation reproduces the parent run step for step. Measured as the vanilla grokking step divided by the rotated one, both counted from initialization, the paired median speedups at $5$k are $1.30$ and $2.19$ for alignments of $0.4$ and $0.7$, and $1.13$ and $1.42$ at $30$k. The rotation down to $0.05$ gives $0.93$ at $5$k and $0.98$ at $30$k. Rotating to $0.7$ shortens the delay on every seed at both checkpoints, least for seed $2$ at $30$k, which had nearly grokked by then.

\paragraph{Matched random rotation.} Twenty vanilla runs, seeds $0$ to $19$, were resumed at $t^{\ast}=5$k from the same saved model and optimizer state without rotation, with a rotation toward the minimal target subspace to an alignment of $0.7$, with a rotation down to an alignment of $0.05$ with it, or with one of two rotations toward a fresh random rank-$16$ subspace. The first random rotation uses the same angle as the rotation toward the target subspace, which moves the embedding by the same Frobenius norm, and the second rotates toward a fresh random rank-$16$ subspace that raises the alignment with that subspace by as much as the rotation toward the target subspace raises its own. Each embedding matrix is rotated separately, and the budget is $300$k steps, with no run censored. With the optimizer state kept, the paired median speedups are $2.40\times$ toward the target subspace, $1.10\times$ toward a random subspace by the same angle, $1.13\times$ toward a random subspace with the same rise, and $0.95\times$ for the rotation down to $0.05$. With the optimizer state reset in every arm, on seeds $0$ to $4$, they are $2.25\times$, $1.08\times$, $1.11\times$ and $1.03\times$. The target subspace is faster than the random rotation of the same angle on all twenty seeds, by $36{,}310$ steps on average, and on all five seeds with the state reset. From the $30$k checkpoint, on the eighteen seeds other than $2$ and $15$, which grok at $32{,}000$ and $27{,}600$ steps, the paired median speedups are $1.35\times$ toward the target subspace, $1.07\times$ toward a random subspace by the same angle, $1.06\times$ with the same rise and $0.99\times$ for the rotation down, and $1.53\times$, $1.09\times$, $1.11\times$ and $1.03\times$ with the optimizer state reset on seeds $0$, $1$, $3$, $4$ and $5$. The target subspace is again faster than the random rotation of the same angle on every seed, with the state kept or reset.

\paragraph{Noise and weight decay.} Gradient noise is injected as Gaussian noise whose standard deviation is $\sigma$ times the root-mean-square of the full gradient, so that training stays full-batch and only the amplitude changes. One sweep sets $\sigma\in\{0,0.1,0.3,1\}$ at weight decay $1$, the other sets the weight decay to $0.25$, $0.5$, $1$ or $2$ without noise. Each setting starts at the chance alignment and at $\alpha_0\in\{0.3,0.6,0.9\}$, with three seeds each and a budget of $300$k steps. The two sweeps share the protocol setting, so that eight sweep points cover seven distinct settings. At the chance alignment, the median grokking step is $137{,}600$, $85{,}700$, $60{,}800$ and $69{,}000$ steps for weight decays of $0.25$, $0.5$, $1$ and $2$ without noise, and $50{,}100$, $49{,}200$ and $45{,}700$ steps for noise levels of $0.1$, $0.3$ and $1$ at weight decay $1$. The slope of $\log\tau$ on $\log\alpha_0$, fitted by least squares to the twelve runs of each setting, lies between $-2.60$ and $-2.42$ in every setting, and none of the seven $95\%$ bootstrap intervals over alignment levels contains zero, the widest being $[-2.89,-1.68]$. Noise and weight decay therefore change the speed of the search, not its dependence on alignment. The restricted network was run under every noise level at weight decay $0.25$ and $2$, three seeds each, and its grokking step stays between $150$ and $280$ steps.

\section{Further Groups}
\label{app:realdata}

The construction applies wherever a group acts visibly on the inputs, and this section reports how far that reaches across groups.

\paragraph{Reach across groups.} The construction is not specific to $S_5$. Table~\ref{tab:groups} collects twelve finite groups of order up to $720$ and gives, for each, the dimension of the minimal target subspace and the vanilla and restricted grokking steps. $Q_8\times C_2$ has no single faithful irrep, and only a faithful set of irreps groks, as the construction predicts. On $Q_8{\times}C_2$ and $\mathrm{PSL}(2,7)$, vanilla training does not grok within its budget of $200$k and $300$k steps, while the restricted network groks at $700$ and $304$ steps, so on these two groups the delay exceeds the budget and restricting the network to the target subspace still removes it.

\paragraph{Further runs on the same groups.} The ceiling holds on these groups as well. Restricted to unfaithful sets whose kernels have $60$ elements on $A_5$, $8$ on $Q_8{\times}C_2$, $24$ and $48$ on $O_h$, $60$ and $120$ on $I_h$, $27$, $54$ and $108$ on $\Sigma(36{\times}3)$ and $360$ on $A_6$, the five seeds of each setting settle on one plateau, between $0.93$ and $1.09$ times the level computed over all input pairs, and no plateau passes the bound on its own test split. On $Q_8{\times}C_2$ the plateau is $21$ of $154$ test pairs against a bound of $26$, and on $A_5$ it is $.0181$ against $.0218$. Ten runs of the input-only commutator and squared-conjugation settings on $\mathrm{PSL}(2,7)$ land exactly on their counts on the test split. Repeats of the restricted settings of Table~\ref{tab:groups}, evaluated every ten steps, grok at $110$ to $120$ steps on $O_h$ and $I_h$ and at $260$ to $340$ on $\Sigma(36{\times}3)$, and the same settings run for $60$k steps end at full test accuracy. Vanilla runs with a budget of $150$k steps grok at $24{,}250$ to $29{,}000$ steps on $O_h$ and at $27{,}000$ to $50{,}000$ on $I_h$. A separate set of $A_6$ runs with a hidden width of $256$, twice that of Table~\ref{tab:groups}, groks at $130$ to $150$ steps restricted and at $11{,}750$ to $19{,}250$ steps vanilla.

\begin{table}[htbp]
\centering\footnotesize
\caption{\textbf{One construction, twelve groups.} Dim.\ is the rank of the projector onto the minimal target subspace. Grokking steps are medians over the runs that grok, and every restricted setting grokked on all of its runs, five per group and ten on $S_5$. Vanilla budgets are $60$k steps, except $170$k for $A_5$, $200$k for $Q_8{\times}C_2$ and $S_5$, $300$k for $\mathrm{PSL}(2,7)$ and $600$k for $\Sigma(36{\times}3)$. At shorter budgets, $\mathrm{PSL}(2,7)$ grokked in $0$ of $5$ vanilla runs at $60$k, and $\Sigma(36{\times}3)$ in $0$ of $5$, $0$ of $3$ and $1$ of $3$ at $60$k, $150$k and $300$k. A sixth $A_5$ seed, $42$, does not grok vanilla within $170$k steps and groks restricted at $125$.}\label{tab:groups}
\setlength{\tabcolsep}{4pt}
\begin{tabular*}{\columnwidth}{@{\extracolsep{\fill}}lrrcrrr@{}}
\toprule
 & & & \multicolumn{2}{c}{Vanilla} & & \\
\cmidrule(lr){4-5}
Group & $|G|$ & Dim. & Runs & $\tau$ & Restr.\ $\tau$ & Speedup \\
\midrule
$\Z_{53}$ & $53$ & $2$ & $5/5$ & $23{,}673$ & $71$ & $333\times$ \\
$\Z_{59}$ & $59$ & $2$ & $5/5$ & $17{,}176$ & $79$ & $217\times$ \\
$\Z_{97}$ & $97$ & $2$ & $5/5$ & $8{,}304$ & $104$ & $80\times$ \\
$Q_8{\times}C_2$ & $16$ & $5$ & $0/5$ & --- & $700$ & --- \\
$A_5$ & $60$ & $9$ & $3/5$ & $109{,}539$ & $126$ & $869\times$ \\
$S_5$ & $120$ & $16$ & $20/20$ & $64{,}050$ & $150$ & $427\times$ \\
$\mathrm{PSL}(2,7)$ & $168$ & $18$ & $0/3$ & --- & $304$ & --- \\
$A_6$ & $360$ & $25$ & $4/5$ & $3{,}323$ & $231$ & $14\times$ \\
$S_6$ & $720$ & $25$ & $5/5$ & $1{,}303$ & $265$ & $5\times$ \\
\midrule
$O_h$ & $48$ & $9$ & $5/5$ & $27{,}840$ & $106$ & $263\times$ \\
$I_h$ & $120$ & $9$ & $5/5$ & $29{,}741$ & $113$ & $263\times$ \\
$\Sigma(36{\times}3)$ & $108$ & $18$ & $3/3$ & $301{,}250$ & $292$ & $1{,}032\times$ \\
\bottomrule
\end{tabular*}
\end{table}

\end{document}